\documentclass[preprint]{elsarticle}

\usepackage{lineno,hyperref}
\modulolinenumbers[5]

\usepackage[utf8]{inputenc}
\usepackage{blindtext}
\usepackage{booktabs}
\usepackage[usenames,dvipsnames,table]{xcolor}
\usepackage{epstopdf}
\usepackage{amsmath}
\usepackage{amssymb}
\usepackage{mathtools}
\usepackage{multirow}
\usepackage{bm}
\usepackage{enumitem}
\usepackage{algorithm}
\usepackage{algcompatible}
\usepackage{xfrac}
\usepackage{adjustbox}
\usepackage{tikz}
\usetikzlibrary{shapes.geometric}
\usepackage{tablefootnote}

\usepackage{pdflscape}
\usepackage{afterpage}

\graphicspath{{}}
\newcommand{\DLTlinesPage}{p.\,180}
\newcommand{\codeURL}{http://www.fit.vutbr.cz/~ipribyl/DLT-based-PnL/}

\algrenewcommand{\alglinenumber}[1]{\footnotesize#1.}
\algrenewcommand{\algorithmiccomment}[1]{\hskip1em // #1}

\newdefinition{rmk}{Remark}
\newcommand{\Alpha}{A}
\newcommand{\Beta}{B}
\newcommand{\Tau}{T}

\newcommand{\rcolorA}{gray!60}
\newcommand{\rcolorB}{gray!45}
\newcommand{\rcolorC}{gray!30}
\let\rtextA\textbf
\let\rtextB\textbf
\let\rtextC\textbf
\newcommand{\rtA}[1]{\colorbox{\rcolorA}{\rtextA{#1}}}
\newcommand{\rtB}[1]{\colorbox{\rcolorB}{\rtextB{#1}}}
\newcommand{\rtC}[1]{\colorbox{\rcolorC}{\rtextC{#1}}}
\newcommand{\rA}[1]{\cellcolor{\rcolorA}\rtextA{#1}}
\newcommand{\rB}[1]{\cellcolor{\rcolorB}\rtextB{#1}}
\newcommand{\rC}[1]{\cellcolor{\rcolorC}\rtextC{#1}}

\definecolor{Ansar}{rgb}{0.2, 0.2, 0.2}
\definecolor{Mirzaei}{rgb}{0.75, 0, 0}
\definecolor{RPnL}{rgb}{0, 0, 0.75}
\definecolor{P3L}{rgb}{0.21176, 0.80392, 1}
\definecolor{ASPnL}{rgb}{0.2549, 0.41176, 0.88235}
\definecolor{LPnLBarLS}{rgb}{0, 0.5451, 0.5451}
\definecolor{LPnLBarENull}{rgb}{0.12549, 0.69804, 0.66667}
\definecolor{DLTlines}{rgb}{0.59, 0.13, 0.59}
\definecolor{DLTplucker}{rgb}{0, 0.75, 0}
\definecolor{DLTcombined}{rgb}{1, 0.69, 0}
\newcommand{\markAnsar}{\begin{tikzpicture}[scale=1]
	\node[
    	draw, regular polygon, regular polygon sides=3,
        minimum size=0em, inner sep=0.13em,
        Ansar, fill=Ansar,
        regular polygon rotate=270
    ] at (0,0) {};
\end{tikzpicture}}
\newcommand{\markMirzaei}{\begin{tikzpicture}
	\draw[Mirzaei,fill=Mirzaei] (0,0) circle (.28em);
\end{tikzpicture}}
\newcommand{\markRPnL}{\begin{tikzpicture}[scale=0.15]
	\draw[RPnL,fill=RPnL,yscale=1.3,rotate=45] (0,0) rectangle (1,1);
\end{tikzpicture}}
\newcommand{\markPthreeL}{\begin{tikzpicture}[scale=0.15]
	\draw[P3L,fill=P3L,yscale=1.3,rotate=45] (0,0) rectangle (1,1);
\end{tikzpicture}}
\newcommand{\markASPnL}{\begin{tikzpicture}[scale=0.15]
	\draw[ASPnL,fill=ASPnL,yscale=1.3,rotate=45] (0,0) rectangle (1,1);
\end{tikzpicture}}
\newcommand{\markLPnLBarLS}{\begin{tikzpicture}[scale=1]
	\node[
    	draw, star, star point ratio=2.5, star points=5,
        minimum size=0em, inner sep=0.1em,
        LPnLBarLS, fill=LPnLBarLS,
        regular polygon rotate=0
    ] at (0,0) {};
\end{tikzpicture}}
\newcommand{\markLPnLBarENull}{\begin{tikzpicture}[scale=1]
	\node[
    	draw, star, star point ratio=2.0, star points=6,
        minimum size=0em, inner sep=0.12em,
        LPnLBarENull, fill=LPnLBarENull,
        regular polygon rotate=0
    ] at (0,0) {};
\end{tikzpicture}}
\newcommand{\markDLTlines}{\begin{tikzpicture}[scale=1]
	\node[
    	draw, regular polygon, regular polygon sides=3,
        minimum size=0em, inner sep=0.13em,
        DLTlines, fill=DLTlines,
        regular polygon rotate=0
    ] at (0,0) {};
\end{tikzpicture}}
\newcommand{\markDLTplucker}{\begin{tikzpicture}[scale=1]
	\node[
    	draw, regular polygon, regular polygon sides=3,
        minimum size=0em, inner sep=0.13em,
        DLTplucker, fill=DLTplucker,
        regular polygon rotate=180
    ] at (0,0) {};
\end{tikzpicture}}
\newcommand{\markDLTcombined}{\begin{tikzpicture}[scale=0.19,baseline=-0.5]
	\draw[DLTcombined,fill=DLTcombined] (0,0) rectangle (1,1);
\end{tikzpicture}}

\journal{Computer Vision and Image Understanding}

\usepackage{numcompress}\biboptions{authoryear}

\begin{document}

\begin{frontmatter}

\title{
	Absolute Pose Estimation from Line Correspondences\\ using Direct Linear Transformation
    \tnoteref{code}
}
\tnotetext[code]{Matlab code and supplementary material are available at \url{\codeURL}.}

\author[addressBUT]{Bronislav Přibyl\corref{correspauthor}}
\cortext[correspauthor]{Corresponding author at BUT FIT, Božetěchova 2, 612 66 Brno, Czech Republic. Phone number +420 777 241 447.}
\ead{ipribyl@fit.vutbr.cz}

\author[addressBUT]{Pavel Zemčík}
\ead{zemcik@fit.vutbr.cz}

\author[addressBUT]{Martin Čadík}
\ead{cadik@fit.vutbr.cz}

\address[addressBUT]{Brno University of Technology,
    Faculty of Information Technology,\\
    Centre of Excellence IT4Innovations,
    Božetěchova 2,
    612 66 Brno,
    Czech Republic
}

\begin{abstract}
This work is concerned with camera pose estimation from correspondences of 3D/2D lines, i.\,e.\ with the Perspective-n-Line (PnL) problem.
We focus on large line sets, which can be efficiently solved by methods using linear formulation of PnL.
We propose a novel method ``DLT-Combined-Lines'' based on the Direct Linear Transformation (DLT) algorithm, which benefits from a new combination of two existing DLT methods for pose estimation.
The method represents 2D structure by lines, and 3D structure by both points and lines.
The redundant 3D information reduces the minimum required line correspondences to 5.
A cornerstone of the method is a combined projection matrix estimated by the DLT algorithm. It contains multiple estimates of camera rotation and translation, which can be recovered after enforcing constraints of the matrix.
Multiplicity of the estimates is exploited to improve the accuracy of the proposed method.
For large line sets (10 and more), the method is comparable to the state-of-the-art in accuracy of orientation estimation. It achieves state-of-the-art accuracy in estimation of camera position and it yields the smallest reprojection error under strong image noise. The method achieves top-3 results on real world data.
The proposed method is also highly computationally effective, estimating the pose of 1000 lines in 12\,ms on a desktop computer.
\end{abstract}


\end{frontmatter}

\section{Introduction}
\label{sec:intro}

Absolute pose estimation is the task of determining the relative position and orientation of a \emph{camera} and an \emph{object} to each other in 3D space. It has many applications in computer vision: 3D reconstruction, robot localization and navigation, visual servoing, and augmented reality are just some of them. The task can be formulated either as object pose estimation (with respect to camera coordinate frame) or as camera pose estimation (with respect to object or world coordinate frame). The latter formulation is used in this paper.

To estimate the camera pose, correspondences between known real world features and their counterparts in the image plane of the camera are needed. The features can be e.\,g.\ points, lines, or combinations of both~\citep{kuang2013pose}. The task has been solved using point correspondences first~\citep{fischler1981random,lowe1987three}. This is called the \emph{Perspective-n-Point} (PnP) problem and it still enjoys attention of researchers~\citep{lepetit2009epnp,ferraz2014very,valeiras2016eventbased}. Camera pose can also be estimated using line correspondences, which is called the \emph{Perspective-n-Line} (PnL) problem. The PnP approach has been studied first, as points are easier to handle mathematically than lines. PnP however is limited only to cases with enough distinctive points, i.\,e.\ mainly to well textured scenes. Conversely, the PnL approach is suitable for texture-less scenes, e.\,g.\ for man-made and indoor environments. Moreover, line features are more stable than point features and are robust to (partial) occlusions.

When estimating camera pose ``from scratch'', the following pipeline is typically used: (i) obtain tentative feature correspondences,  (ii) filter out outliers, (iii) compute a solution from all inliers, and (iv), optionally, iteratively refine the solution, e.\,g.\ by minimizing reprojection error. Task (ii) is usually carried out by iterative solving of a problem with a minimal number of line correspondences (i.\,e.\ P3L) in a RANSAC loop. Task (iii), on the other hand, requires solving a problem with high number of lines.
In some applications, the correspondences are already known and thus only task (iii) is to be solved.
 
In recent years, versatile PnL methods have been developed which are suitable for both of these tasks. Remarkable progress has been achieved \citep{ansar2003linear,mirzaei2011globally,zhang2012rpnl}, mainly in accuracy of the methods, in their robustness to image noise, and in their effectiveness. These methods are outperformed in task (iii) however, by LPnL methods -- methods based on a linear formulation of the PnL problem \citep{Pribyl2015,xu2016pnl}.
LPnL methods are superior in terms of both accuracy and computational speed in camera pose estimation from many ({\footnotesize $\sim$}~tens to thousands) line correspondences. The oldest LPnL method is that proposed by \citet[\DLTlinesPage]{hartley2004multiple}, followed recently by the method of \cite{Pribyl2015}.
Even more recently,
\cite{xu2016pnl} introduced a series of LPnL methods generated by the use of Cartesian or barycentric coordinates, and by alternating whether the solution is retrieved in closed form or by optimization.
As we show in this paper, space for improving accuracy of the methods still exists.

In this paper, we introduce a novel method based on linear formulation of the PnL problem, which is a combination of the DLT-Lines method of \cite{hartley2004multiple} and the DLT-Plücker-Lines method of \cite{Pribyl2015}. The former represents the 3D structure by 3D points, while the latter represents it by 3D lines parameterized by Plücker coordinates.
The proposed method exploits the redundant representation of 3D structure by both 3D points and 3D lines, which leads to the reduction of the minimum required line correspondences to~5.
A cornerstone of the method is a combined projection matrix recovered by the DLT algorithm. It contains multiple estimates of camera orientation and translation, enabling a more accurate estimation of the final camera pose.
The proposed method achieves state-of-the-art accuracy for large line sets under strong image noise, and it performs comparably to state-of-the-art methods on real world data.
The proposed method also keeps the common advantage of LPnL methods -- being very fast.

The rest of this paper is organized as follows.
We present a review of related work on PnL in Section~\ref{sec:related-work}.
Then we introduce mathematical notation and Plücker coordinates of 3D lines, and show how points and lines transform and project onto the image plane in Section~\ref{sec:transformations}.
In Section~\ref{sec:DLT}, we explain the application of DLT algorithm to the PnL problem in general, and we describe the existing methods DLT-Lines and DLT-Plücker-Lines.
In Section~\ref{sec:DLT-Combined-Lines}, we propose the novel method DLT-Combined-Lines.
We evaluate the performance of the proposed method using simulations and real-world experiments in Section~\ref{sec:results},
and we conclude in Section~\ref{sec:conclusions}.

\section{Related work}
\label{sec:related-work}

The task of camera pose estimation from line correspondences has been receiving attention for more than a quarter of century. Some of the earliest works are those by \cite{dhome1989determination} and \cite{liu1990determination}. They introduce two different ways to deal with the PnL problem  -- algebraic and iterative approaches -- both of which have different properties and thus also different uses. A specific subset of algebraic approaches are the methods based on linear formulation of the PnL problem.

\subsection{Iterative methods}

The iterative approaches consider pose estimation as a nonlinear least squares problem by iteratively minimizing specific error function, which usually has a geometrical meaning. In the early work of \cite{liu1990determination}, the authors attempted to estimate the camera position and orientation separately developing a method called R\_then\_T.
Later on, \cite{kumar1994robust} introduced a method called R\_and\_T for simultaneous estimation of camera position and orientation, and proved its superior performance to R\_then\_T.
Recently, \cite{zhang2016probabilistic} proposed two modifications of the R\_and\_T algorithm exploiting the uncertainty properties of line segment endpoints.  
Several other iterative methods are also capable of \emph{simultaneous} estimation of pose parameters and line correspondences, e.\,g.~\cite{david2003simultaneous,zhang2012unknown}. They pose an orthogonal approach to the common RANSAC-based correspondence filtering and consecutive separate pose estimation.

Iterative algorithms suffer from two common major issues when not initialized accurately: They converge slowly, and more severely, the estimated pose is often far from the true camera pose, finding only a local minimum of the error function.
This makes iterative approaches suitable for final refinement of an initial solution, provided by some other algorithm.

\subsection{Algebraic methods}

The algebraic approaches estimate the camera pose by solving a system of (usually polynomial) equations, minimizing an algebraic error. Their solutions are thus not necessarily geometrically optimal; on the other hand, no initialization is needed.

Among the earliest efforts in this field are those of \cite{dhome1989determination} and \cite{chen1990pose}. Both methods solve the minimal problem of pose estimation from 3 line correspondences in a closed form.
Solutions of the P3L problem are multiple: up to 8 solutions may exist \citep{chen1990pose}.
Unfortunately, neither method is able to exploit more measurements to remove the ambiguity, and both methods are sensitive to presence of image noise.

\textbf{Ansar} and Daniilidis (\citeyear{ansar2003linear}) developed a method that is able to handle 4 or more lines, limiting the number of possible solutions to 1.
Lifting is employed to convert  a polynomial system to linear equations in the entries of a rotation matrix. This approach may, however, fail in cases of singular line configurations (e.\,g.\ lines in 3 orthogonal directions -- \citeauthor{navab1993critical}, \citeyear{navab1993critical}) as the underlying polynomial system may have multiple solutions.
The algorithm has quadratic computational complexity ($O(n^2)$, where $n$ is the number of lines), which renders it impractically slow for processing higher numbers of lines. The method also becomes unstable with increasing image noise, eventually producing solutions with complex numbers.

Recently, two major improvements of algebraic approaches have been achieved. First, \textbf{Mirzaei} and Roumeliotis (\citeyear{mirzaei2011globally}) proposed a method, which is more computationally efficient ($O(n)$), behaves more robustly in the presence of image noise, and can handle the minimum of 3 lines, or more.
A polynomial system with 27 candidate solutions is constructed and solved through the eigendecomposition of a multiplication matrix.
Camera orientations having the least square error are considered to be the optimal ones. Camera positions are obtained separately using linear least squares. A weakness with this algorithm is that it often yields multiple solutions.
Also, despite its linear computational complexity, the overall computational time is still high due to slow construction of the
multiplication matrix, which causes a high constant time penalty: 78\,ms\,/\,10 lines.

The second recent improvement is the Robust PnL (\textbf{RPnL}) algorithm of \cite{zhang2012rpnl}. Their method works with 4 or more lines and is more accurate and robust than the method of Mirzaei and Roumeliotis. An intermediate model coordinate system is used in the method of Zhang et al., which is aligned with a 3D line of longest projection. The lines are divided into triples, for each of which a P3L polynomial is formed. The optimal solution of the polynomial system is selected from the roots of its derivative in terms of a least squares residual.

The RPnL algorithm was later modified by \cite{xu2016pnl} into the Accurate Subset based PnL (\textbf{ASPnL}) algorithm, which acts more accurately on small line sets. However, it is very sensitive to outliers, limiting its performance on real-world data. This algorithm is compared to other state-of-the-art methods in Section~\ref{sec:results}. A drawback of both RPnL and ASPnL is that their computational time increases strongly for higher number of lines -- from 8\,ms\,/\,10 lines to 630 -- 880\,ms\,/\,1000 lines.

\subsection{Methods based on linear formulation of PnL}

A specific subset of algebraic methods are methods exploiting a linear formulation of the PnL problem (LPnL). Generally, the methods solve a system of linear equations, the size of which is directly proportional to the number of measurements. The biggest advantage of LPnL methods is their computational efficiency, making them fast for both low and high number of lines.

The most straightforward way to solve LPnL is the Direct Linear Transformation (DLT) algorithm \citep{hartley2004multiple}. It transforms the measured line correspondences into a homogeneous system of linear equations, whose coefficients are arranged into a measurement matrix. The solution then lies in the nullspace of the matrix. A necessary condition to apply any DLT method on noisy data is to prenormalize the input in order to ensure that the entries of the measurement matrix are of equal magnitude. Otherwise, the method will be oversensitive to noise and it will produce results arbitrarily far from the true solution.

The first DLT method for solving PnL is the method of \citet[\DLTlinesPage]{hartley2004multiple}. Following the terminology of~\cite{silva2012cameraDLT}, we call the method \textbf{DLT-Lines}.
It does not act directly on 3D lines, but rather on 3D \emph{points} lying on 3D lines (for example line endpoints).
It exploits the fact that if a 3D line and a 3D point coincide, their projections also must coincide.
The DLT-Lines method requires at least 6 line correspondences.

Recently, \cite{Pribyl2015} developed a DLT method, which acts on 3D lines directly. The lines are parameterized using Plücker coordinates, hence the name of the method is \textbf{DLT-Plücker-Lines}. The method yields more accurate estimates of camera orientation than DLT-Lines at the cost of a bit larger reprojection error and slightly lower computational efficiency. Also, the minimum number of lines required is 9.

Even more recently, \cite{xu2016pnl} introduced a new set of methods exploiting the linear formulation of the PnL problem. The authors were inspired by the state-of-the-art PnP solver working on the same principle \citep{ferraz2014very}. Similarly to DLT-Lines, the new methods act on 3D \emph{points} and 2D lines. The methods of \cite{xu2016pnl} can be categorized by two criteria.
Firstly, by parameterization of 3D points (either by Cartesian or by barycentric coordinates -- this is denoted in the method's names by \emph{DLT} and \emph{Bar}, respectively).
Secondly, by the manner in which a solution is obtained from the nullspace. The solution is either an exact rank-1 nullspace computed in closed form using homogeneous linear least squares\footnote{
	We use the term ``homogeneous linear least squares'' to denote solving of a homogeneous linear system $\mathsf{M}\mathbf{p} = \mathbf{0}$ for $\mathbf{p}$ which is done by minimization of $||\mathsf{M}\mathbf{p}||$ subject to $||\mathbf{p}||=1$. The correct notation, however somewhat confusing, would be a ``low-rank approximation'' (of $\mathsf{M}$).
}, or it is estimated from an ``effective nullspace'' \citep{lepetit2009epnp} of a dimension 1 -- 4 (higher dimensions typically occurring under the presence of noise). This is denoted in the method's names by \emph{LS} and \emph{ENull}, respectively. All the following methods require at least 6 line correspondences, although the effective null space solver (ENull) is sometimes able to recover the correct solution of an underdetermined system defined by 4 or 5 lines.
The four LPnL methods of Xu et al. are the following:

\noindent\textbf{LPnL\-\_DLT\-\_LS} parameterizes 3D points using Cartesian coordinates, and it uses homogeneous linear least squares to recover the solution: entries of the rotation matrix and translation vector. This is exactly the same algorithm as DLT-Lines \cite[\DLTlinesPage]{hartley2004multiple}, so we use the name \emph{DLT-Lines} to refer to the method in the rest of the paper.

\noindent\textbf{LPnL\-\_DLT\-\_ENull} parameterizes 3D points using Cartesian coordinates, and it uses the effective nullspace solver \citep{lepetit2009epnp} to recover the solution: entries of the rotation matrix and translation vector.
It achieves higher accuracy than DLT-Lines.

\noindent\textbf{LPnL\-\_Bar\-\_LS} parameterizes 3D points using barycentric coordinates, which depend on the position of 4 arbitrarily chosen control points. Position of the control points with respect to camera is solved using homogeneous linear least squares. Alignment of the 4 camera- and world-referred control points defines the camera pose. Accuracy of the method is similar to DLT-Lines.

\noindent\textbf{LPnL\-\_Bar\-\_ENull} parameterizes 3D points using barycentric coordinates. Position of the 4 control points with respect to camera is solved using the effective nullspace solver. Alignment of the 4 camera- and world-referred control points defines the camera pose.
The method is even more accurate than LPnL\-\_Bar\-\_LS.

In this paper, we exploit the common properties of DLT-Lines and DLT-Plücker-Lines methods and we combine them into a new method \textbf{DLT-Com\-bined-Lines}.
As a result, the minimal number of line correspondences required by the proposed method is reduced to 5, position of the camera is estimated more accurately under strong image noise than by the existing most accurate method (LPnL\-\_Bar\-\_ENull), and the method yields lower reprojection error. Accuracy of orientation estimates is similar to the state-of-the-art method.
The proposed method also benefits from the common advantage of all LPnL methods -- being very fast.

\section{Transformations of points and lines}
\label{sec:transformations}

In this section, we introduce notation, define coordinate systems, and also
define parameterization of 3D lines using Plücker coordinates. Then, we review how points and lines are transformed in Euclidean space, and how they project onto the image plane using central projection.

\subsection{Notation and coordinate systems}
\label{subsec:notation}

Scalars are typeset in italics ($x, X$), vectors are typeset in bold ($\mathbf{l}$, $\mathbf{L}$).
All vectors are thought of as being column vectors unless explicitly transposed.
Matrices are typeset in sans-serif fonts ($\mathsf{t}$, $\mathsf{D}$), the identity matrix is denoted by $\mathsf{I}$.
2D entities are denoted by lower case letters ($x$, $\mathbf{l}$, $\mathsf{t}$), 3D entities by upper case letters ($X$, $\mathbf{L}$, $\mathsf{D}$).
No formal distinction between coordinate vectors and physical entities is made.
Equality of up to a non-zero scale factor is denoted by $\approx$, transposition by ${}^\top$, Euclidean norm of a vector by $||.||$, Kronecker product by $\otimes$, vectorization of a matrix in column-major order by ``$\mathrm{vec}(.)$'', and the skew symmetric $3 \times 3$ matrix associated with the cross product by $[.]_{\times}$, i.\,e.\ $[\mathbf{a}]_{\times}\mathbf{b} = \mathbf{a} \times \mathbf{b}$.
Transformation matrices acting on points and lines are distinguished by a dot and a bar, respectively ($\dot{\mathsf{D}}$, $\bar{\mathsf{D}}$).

Let us now define the coordinate systems: a world coordinate system and a camera coordinate system. Both systems are right-handed. The camera $X$-axis goes right, the $Y$-axis goes up and the $Z$-axis goes behind the camera, so that the points placed in front of the camera have negative $Z$ coordinates in the camera coordinate system. A transition from the world to the camera coordinate system is realized through a translation followed by a rotation. The translation is parameterized using a $3 \times 1$ translation vector $\mathbf{T} = (T_1 ~ T_2 ~ T_3)^\top$, which represents the position of the camera in the world coordinate system. The rotation is parameterized using a $3 \times 3$ rotation matrix $\mathsf{R}$ describing the orientation of the camera in the world coordinate system by means of three consecutive rotations along the three axes $Z$, $Y$, $X$ by respective Euler angles $\Gamma$, $\Beta$, $\Alpha$. Camera pose is thus parameterized by $T_1$, $T_2$, $T_3$, $\Alpha$, $\Beta$, $\Gamma$.

In the following sections, a pinhole camera with known intrinsic parameters is assumed.

\subsection{Transformation of a point}
\label{subsec:transfpoint}

A homogeneous 3D point $\mathbf{X} = (X_1 ~ X_2 ~ X_3 ~ X_4)^\top$ in the world coordinate system is transformed to a point $\dot{\mathsf{D}} \mathbf{X}$ in the camera coordinate system using a $4 \times 4$ point displacement matrix $\dot{\mathsf{D}}$, where
\begin{equation}
	\dot{\mathsf{D}} \approx
	\left[ \begin{array}{lc}
		\mathsf{R} & -\mathsf{R} \mathbf{T} \\
		\mathbf{0}_{\scriptscriptstyle 1 \times 3} & 1 
	\end{array} \right] \enspace .
  \label{eq:pointdisplacementmatrix}
\end{equation}

\noindent After 3D points are transformed into the camera coordinate system, they can be projected onto the normalized image plane using the $3 \times 4$ canonical camera matrix $(\mathsf{I} \enspace \mathbf{0})$. Compositing the two transformations yields the $3 \times 4$ point projection matrix
\begin{equation}
  \dot{\mathsf{P}} \approx
	\left[ \begin{array}{ccc}
		\mathsf{R} & & -\mathsf{R}\mathbf{T} 
	\end{array} \right] \enspace .
  \label{eq:pointprojectionmatrix}
\end{equation}

\noindent A 3D point ${\mathbf{X}}$ is then projected using the point projection matrix $\dot{\mathsf{P}}$ as
\begin{equation}
  {\mathbf{x}} \approx \dot{\mathsf{P}} {\mathbf{X}} \enspace ,
  \label{eq:pointprojection}
\end{equation}

\noindent where ${\mathbf{x}} = (x_1 ~ x_2 ~ x_3)^\top$ is a homogeneous 2D point in the normalized image plane.

\subsection{Plücker coordinates of 3D lines}
\label{subsec:pluckercoords}

3D lines can be represented using several parameterizations in the projective space \citep{bartoli2005structure}. Parameterization using Plücker coordinates is complete (i.\,e.\ every 3D line can be represented) but not minimal (a 3D line has 4 degrees of freedom but Plücker coordinate is a homogeneous 6-vector). The benefit of using Plücker coordinates is in convenient linear projection of 3D lines onto the image plane.

Given two distinct 3D points $\mathbf{X} = (X_1 ~ X_2 ~ X_3 ~ X_4)^\top$ and $\mathbf{Y} = (Y_1 ~ Y_2 ~ Y_3 ~ Y_4)^\top$ in homogeneous coordinates, a line joining them can be represented using Plücker coordinates as a homogeneous 6-vector $\mathbf{L} \approx (\mathbf{U}^\top ~ \mathbf{V}^\top)^\top = (L_1 ~ L_2 ~ L_3 ~ L_4 ~ L_5 ~ L_6)^\top$, where
\begin{eqnarray}
	\label{eq:plucker}
	\mathbf{U}^\top &=& (L_1 ~ L_2 ~ L_3) = (X_1 ~ X_2 ~ X_3) ~ \times ~ (Y_1 ~ Y_2 ~ Y_3) \enspace , \\ \nonumber
	\mathbf{V}^\top &=& (L_4 ~ L_5 ~ L_6) = X_4(Y_1 ~ Y_2 ~ Y_3) ~ - ~ Y_4(X_1 ~ X_2 ~ X_3) \enspace .
\end{eqnarray}

\noindent The $\mathbf{V}$ part encodes direction of the line while the $\mathbf{U}$ part encodes position of the line in space. In fact, $\mathbf{U}$ is a normal of an interpretation plane -- a plane passing through the line and the origin. As a consequence, $\mathbf{L}$ must satisfy a bilinear constraint $\mathbf{U}^\top \mathbf{V} = 0$. Existence of this constraint explains the discrepancy between the 4 degrees of freedom of a 3D line and its parameterization by a homogeneous 6-vector. More on Plücker coordinates can be found in~\cite{hartley2004multiple}.

\subsection{Transformation of a line}
\label{subsec:transfline}

A 3D line parameterized using Plücker coordinates can be transformed from the world into the camera coordinate system using the $6 \times 6$ line displacement matrix $\bar{\mathsf{D}}$ \citep{bartoli20043d}\footnote{
	Please note that our line displacement matrix differs slightly from the matrix of \citet[Eq.~(6)]{bartoli20043d}, namely in the upper right term: We have $\mathsf{R} [-\mathbf{T}]_{\times}$ instead of $[\mathbf{T}]_{\times} \mathsf{R}$ due to our different coordinate system.
}, where
\begin{equation}
    \bar{\mathsf{D}} \approx
	\left[ \begin{array}{lc}
		\mathsf{R} & \mathsf{R} [-\mathbf{T}]_{\times} \\
		\mathsf{0}_{\scriptscriptstyle 3 \times 3} & \mathsf{R} 
	\end{array} \right] \enspace .
  \label{eq:linedisplacementmatrix}
\end{equation}

\begin{figure}
	\centering
	\includegraphics[width=0.45\linewidth]{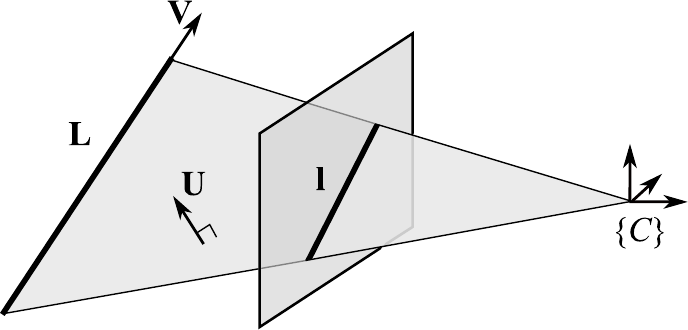}
    \caption{
		3D line projection. The 3D line $\mathbf{L}$ is parameterized by its direction vector $\mathbf{V}$ and a normal $\mathbf{U}$ of its interpretation plane, which passes through the origin of the camera coordinate system $\{C\}$. Since the projected 2D line $\mathbf{l}$ lies at the intersection of the interpretation plane and the image plane, it is fully defined by the normal $\mathbf{U}$.
	}
    \label{fig:projection}
\end{figure}

\noindent After 3D lines are transformed into the camera coordinate system, their projections onto the image plane can be determined as intersections of their interpretation planes with the image plane; see Fig.~\ref{fig:projection} for illustration.
The normal $\mathbf{U}$ of an interpretation plane is identical to the image line $\mathbf{l}$ in the coordinate system of the camera, hence only $\mathbf{U}$ needs to be computed when projecting $\mathbf{L}$, and only the upper half of $\bar{\mathsf{D}}$ is needed, yielding the $3 \times 6$ line projection matrix \citep{faugeras1995geometry}
\begin{equation}
  \bar{\mathsf{P}} \approx
	\left[ \begin{array}{ccc}
		\mathsf{R} & & \mathsf{R} [-\mathbf{T}]_{\times} 
	\end{array} \right] \enspace .
  \label{eq:lineprojectionmatrix}
\end{equation}

\noindent A 3D line ${\mathbf{L}}$ is then projected using the line projection matrix $\bar{\mathsf{P}}$ as
\begin{equation}
  {\mathbf{l}} \approx \bar{\mathsf{P}} {\mathbf{L}} \enspace ,
  \label{eq:lineprojection}
\end{equation}

\noindent where $\mathbf{l} = (l_1 ~ l_2 ~ l_3)^\top$ is a homogeneous 2D line in the normalized image plane.

\section{Pose estimation using DLT}
\label{sec:DLT}

We will first describe DLT methods in general in Section~\ref{subsec:DLT-general} and show the common steps. Then, we will describe the DLT-Lines method in Section~\ref{subsec:DLT-Lines} and the DLT-Plücker-Lines method in Section~\ref{subsec:DLT-Plucker-Lines}. Finally, we will briefly describe an algebraic scheme to deal with outlying line correspondences in Section~\ref{subsec:AOR}.

\subsection{General structure of DLT}
\label{subsec:DLT-general}

Let us assume that we have (i) a calibrated pinhole camera and (ii) correspondences between 3D lines (or 3D points lying on those lines) and images of the lines obtained by the camera. Given these requirements, it is possible to estimate the camera pose using a DLT method. The methods have the following steps in common:
\begin{enumerate}
	\item Input data are prenormalized to achieve good conditioning of the linear system.
    \item A projection matrix is estimated using homogeneous linear least squares, and the effect of prenormalization is reverted.
\item The pose parameters are extracted from the estimated projection matrix. This also includes constraint enforcement in the case of noisy data, since the constraints are not taken into account during the least squares estimation.
\end{enumerate}

\paragraph{Prenormalization}
Since the DLT algorithm is sensitive to the choice of coordinate system, it is crucial to prenormalize the data to get a properly conditioned measurement matrix $\mathsf{M}$ \citep{hartley1997defense}. Various transformations can be used, but the optimal ones are unknown.
In practice, however, the goal is to reduce large absolute values of point/line coordinates. This is usually achieved by centering the data around the origin and by scaling them so that an average coordinate has the absolute value of 1.

Specific prenormalizing transformations are proposed for each method in the following sections.

\paragraph{Linear estimation of a projection matrix}
As a starting point, a system of linear equations needs to be constructed, which relates (prenormalized) 3D entities with their (prenormalized) image counterparts through a projection matrix, denoted $\mathsf{P}$.
This might be
the projection of homogeneous 3D points ${\mathbf{x}} \approx \dot{\mathsf{P}} {\mathbf{X}}$ in Eq.~(\ref{eq:pointprojection}),
or the projection of Plücker lines ${\mathbf{l}} \approx \bar{\mathsf{P}} {\mathbf{L}}$ in Eq.~(\ref{eq:lineprojection}),
or any other linear system, or some combination of these.
The problem of camera pose estimation now resides in estimating the projection matrix $\mathsf{P}$, which encodes all the six camera pose parameters $T_1$, $T_2$, $T_3$, $\Alpha$, $\Beta$, $\Gamma$.

The system of linear equations is transformed so that only a zero vector remains at the right hand side (see \ref{sec:appendix-corresp} for details). The transformed system can be written in the form
\begin{equation}
	\label{eq:system}
	\mathsf{M} \mathbf{p} = \mathbf{0} \enspace ,
\end{equation}

\noindent where $\mathsf{M}$ is a measurement matrix containing coefficients of equations generated by correspondences between 3D entities and their image counterparts.
Each of the $n$ correspondences gives rise to a number of independent linear equations (usually 2), and thus to the same number of rows of $\mathsf{M}$. The number of columns of $\mathsf{M}$ equals $d$, which is the number of entries contained in $\mathsf{P}$. The size of $\mathsf{M}$ is thus $2n \times d$.
Eq.~(\ref{eq:system}) is then solved for the $d$-vector $\mathbf{p} = \mathrm{vec}(\mathsf{P})$.

Eq.~(\ref{eq:system}), however, holds only in the noise-free case. If a noise is present in the measurements, an inconsistent system is obtained:
\begin{equation}
	\label{eq:noisysystem}
	\mathsf{M} \mathbf{p}' = \boldsymbol{\epsilon} \enspace .
\end{equation}

\noindent Only an approximate solution $\mathbf{p}'$ may be found through minimization of a $2n$-vector of measurement residuals $\boldsymbol{\epsilon}$ in the least squares sense,
subject to $||\mathbf{p}'||=1$.

Once the system of linear equations given by (\ref{eq:noisysystem}) is solved,
e.\,g.\ by Singular Value Decomposition (SVD) of $\mathsf{M}$, the estimate $\mathsf{P}'$ of the projection matrix $\mathsf{P}$ can be recovered from the $d$-vector $\mathbf{p}'$.

\paragraph{Extraction of pose parameters}
The estimate $\mathsf{P}'$ of a projection matrix $\mathsf{P}$ obtained as a solution of (\ref{eq:noisysystem})
does not satisfy the constraints imposed on $\mathsf{P}$. In fact, $\mathsf{P}$ has only 6 degrees of freedom -- the 6 camera pose parameters $T_1$, $T_2$, $T_3$, $\Alpha$, $\Beta$, $\Gamma$. $\mathsf{P}$ has, however, more entries: The $3 \times 4$ point projection matrix $\dot{\mathsf{P}}$ has 12 entries and the $3 \times 6$ line projection matrix $\bar{\mathsf{P}}$ has 18 entries. This means that the projection matrices have 6 and 12 independent linear constraints, respectively, see Eq.~(\ref{eq:pointprojectionmatrix}, \ref{eq:lineprojectionmatrix}). The first six constraints are imposed by the rotation matrix $\mathsf{R}$ that must satisfy the orthonormality constraints (unit-norm and mutually orthogonal rows). The other six constraints in the case of $\bar{\mathsf{P}}$ are imposed by the skew-symmetric matrix $[-\mathbf{T}]_{\times}$ (three zeros on the main diagonal and antisymmetric off-diagonal elements).

In order to extract the pose parameters, the scale of $\mathsf{P}'$ has to be corrected first, since $\mathbf{p}'$ is usually of unit length as a minimizer of $\boldsymbol{\epsilon}$ in Eq.~(\ref{eq:noisysystem}).
The correct scale of $\mathsf{P}'$ can only be determined from the part which does not contain the translation $\mathbf{T}$. In both cases of $\dot{\mathsf{P}}$ and $\bar{\mathsf{P}}$, it is the left $3 \times 3$ submatrix $\mathsf{P}'_1$ -- an estimate of a rotation matrix $\mathsf{R}$.
We recommend a method of scale correction based on the fact that all three singular values of a proper rotation matrix should be 1, see Algorithm~\ref{alg:scale}.
Alternatively, the scale can also be corrected so that $\det (s\mathsf{P}'_1) = 1$, but Algorithm~\ref{alg:scale} proved more robust in practice.

\begin{algorithm}
	\normalsize
	\caption{Scale correction of a projection matrix.}
	\label{alg:scale}
	\begin{algorithmic}[1]
    	\REQUIRE An estimate $\mathsf{P}'$ of a projection matrix, possibly wrongly scaled and without the constraints being fulfilled.
        \STATE $\mathsf{P}'_1 \leftarrow$ left $3 \times 3$ submatrix of $\mathsf{P}'$
    	\STATE $\mathsf{U \Sigma V}^\top \leftarrow \mathrm{SVD}(\mathsf{P}'_1)$
        \STATE $s \leftarrow {1}/{\mathrm{mean}(\mathrm{diag}(\mathsf{\Sigma}))}$
        \ENSURE $s \mathsf{P}'$.
    \end{algorithmic}
\end{algorithm}

Further steps in the extraction of pose parameters differ in each method, they are thus part of the description of each method in the following sections.

\subsection{DLT-Lines}
\label{subsec:DLT-Lines}

This is the method introduced by \citet[\DLTlinesPage]{hartley2004multiple}. It exploits the fact that a 3D point $\mathbf{X}$ lying on a 3D line $\mathbf{L}$ projects such that its projection $\mathbf{x} = \dot{\mathsf{P}} \mathbf{X}$ must also lie on the projected line: $\mathbf{l}^\top \mathbf{x} = 0$. Putting this together yields the constraint equation
\begin{equation}
	\label{eq:DLT-Lines}
	\mathbf{l}^\top \dot{\mathsf{P}} \mathbf{X} = 0 \enspace .
\end{equation}

\noindent The pose parameters are encoded in the $3 \times 4$ point projection matrix $\dot{\mathsf{P}}$, see Eq.~(\ref{eq:pointprojectionmatrix}). Since $\dot{\mathsf{P}}$ has 12 entries, at least 6 lines are required to fully determine the system, each line with 2 or more points on it.

\paragraph{Prenormalization}
The known quantities of Eq.~(\ref{eq:DLT-Lines}), i.\,e.\ the coordinates of 3D points and 2D lines, need to be prenormalized.
In the case of the DLT-based pose estimation from points, \cite{hartley1998minimizing} suggests to translate and scale both 3D and 2D points so that their centroid is at the origin and their average distance from the origin is $\sqrt{3}$ and $\sqrt{2}$, respectively.
By exploiting the principle of duality \citep{coxeter2003projective}, we suggest treating coordinates of 2D lines as homogeneous coordinates of 2D points, and then following Hartley in the prenormalization procedure -- i.\,e.\ to apply translation and anisotropic scaling.

\paragraph{Linear estimation of the point projection matrix}
The point projection matrix $\dot{\mathsf{P}}$ and its estimate $\dot{\mathsf{P}}'$ are $3 \times 4$, so the corresponding measurement matrix $\dot{\mathsf{M}}$ is $n \times 12$, where $n$ is the number of point-line correspondences $\mathbf{X}_i \leftrightarrow \mathbf{l}_i$, $(i = 1 \dots n, ~ n \ge 11)$.
$\dot{\mathsf{M}}$ is constructed as
\begin{equation}
	\label{eq:pointmeasurementmatrix}
    \dot{\mathsf{M}}_{\scriptscriptstyle (i,~:)} = \mathbf{X}_i^\top \otimes \mathbf{l}_i^\top \enspace ,
\end{equation}

\noindent where $\dot{\mathsf{M}}{\scriptscriptstyle (i,~:)}$ denotes the $i$-th row of $\dot{\mathsf{M}}$ in Matlab notation.
See \ref{subsec:appendix-corresp-point-line} for a derivation of Eq.~(\ref{eq:pointmeasurementmatrix}).
The 3D points $\mathbf{X}_i$ must be located on at least 6 different lines.

\paragraph{Extraction of pose parameters}
First, the scale of $\dot{\mathsf{P}}'$ is corrected using Algorithm~\ref{alg:scale}, yielding $s \dot{\mathsf{P}}'$.
Then, the left $3 \times 3$ submatrix of $s \dot{\mathsf{P}}'$ is taken as the estimate $\mathsf{R}'$ of a rotation matrix. 
A nearest rotation matrix $\mathsf{R}$ is found in the sense of the Frobenius norm using Algorithm~\ref{alg:orthogonalization}. 

\begin{algorithm}
	\normalsize
	\caption{Orthogonalization of a $3 \times 3$ matrix.}
	\label{alg:orthogonalization}
	\begin{algorithmic}[1]
    	\REQUIRE A $3 \times 3$ estimate $\mathsf{R}'$ of a rotation matrix $\mathsf{R}$.
    	\STATE $\mathsf{U \Sigma V}^\top \leftarrow \mathrm{SVD}(\mathsf{R}')$
        \STATE $d \leftarrow \det (\mathsf{U} \mathsf{V}^\top)$
        \STATE $\mathsf{R} \leftarrow d \mathsf{U} \mathsf{V}^\top$
        \ENSURE $\mathsf{R}$.
    \end{algorithmic}
\end{algorithm}

\noindent Please note that Algorithms \ref{alg:scale} and \ref{alg:orthogonalization} can be combined and executed at once.
The remaining pose parameter to recover is the translation vector $\mathbf{T}$, which is encoded in the fourth column $\dot{\mathbf{P}}_4'$ of $\dot{\mathsf{P}}'$, see Eq.~(\ref{eq:pointprojectionmatrix}).
It is recovered as $\mathbf{T} = s \mathsf{R}^\top \dot{\mathbf{P}}_4'$, completing the extraction of pose parameters.

\subsection{DLT-Plücker-Lines}
\label{subsec:DLT-Plucker-Lines}

This is the method introduced by \cite{Pribyl2015}. It exploits the linear projection of 3D lines parameterized using Plücker coordinates onto the image plane, as described in Section~\ref{subsec:pluckercoords}.
The constraint equation defines the formation of 2D lines $\mathbf{l}$ as projections of 3D lines $\mathbf{L}$, as defined in Eq.~(\ref{eq:lineprojection}):
\begin{equation}
  \mathbf{l} \approx \bar{\mathsf{P}} \mathbf{L} \enspace .
  \label{eq:DLT-Plucker-Lines}
\end{equation}

\noindent The pose parameters are encoded in the $3 \times 6$ line projection matrix $\bar{\mathsf{P}}$, see Eq.~(\ref{eq:lineprojectionmatrix}).
Since $\bar{\mathsf{P}}$ has 18 entries, at least 9 lines are required to fully determine the system.

\paragraph{Prenormalization}
The known quantities of Eq.~(\ref{eq:DLT-Plucker-Lines}), i.\,e.\ the Plücker coordinates of 3D lines, and the coordinates of 2D lines, need to be prenormalized.
Since the homogeneous Plücker coordinates of a 3D line $\mathbf{L}$ cannot be treated as homogeneous coordinates of a 5D point (because of the bilinear constraint, see Section~\ref{subsec:pluckercoords}), we suggest the following prenormalization:
Translation and scaling, which can be applied through the line similarity matrix \citep{bartoli20043d}, affects only the $\mathbf{U}$ part of $\mathbf{L}$. Therefore, the $\mathbf{V}$ parts are adjusted first by multiplying each $\mathbf{L}$ by a non-zero scale factor so that $||\mathbf{V}|| = \sqrt{3}$.
Then, translation is applied to minimize the average magnitude of $\mathbf{U}$. Since $||\mathbf{U}||$ decreases with the distance of $\mathbf{L}$ from the origin, it is feasible to translate the lines so that the sum of squared distances from the origin is minimized. This can be efficiently computed using the Generalized Weiszfeld algorithm \citep{aftab1lqclosest}.
Finally, anisotropic scaling is applied so that the average magnitude of $\mathbf{U}$ matches the average magnitude of $\mathbf{V}$.

Prenormalization of 2D lines can be carried out in the same way as in the case of the DLT-Lines method, see Section~\ref{subsec:DLT-Lines}.

\paragraph{Linear estimation of the line projection matrix}
The line projection matrix $\bar{\mathsf{P}}$ and its estimate $\bar{\mathsf{P}}'$ are $3 \times 6$, so the corresponding measurement matrix $\bar{\mathsf{M}}$ has 18 columns.
The number of its rows depends on the number $m$ of line-line correspondences $\mathbf{L}_j \leftrightarrow \mathbf{l}_j$, $(j = 1 \dots m, ~ m \ge 9)$.
By exploiting Eq.~(\ref{eq:DLT-Plucker-Lines}), each correspondence generates three rows of $\bar{\mathsf{M}}$ (Matlab notation is used to index the matrix elements):
\begin{equation}
	\label{eq:linemeasurementmatrix}
	\bar{\mathsf{M}}_{\scriptscriptstyle (3j-2 \, : \, 3j,~:)} = \mathbf{L}_j^\top ~ \otimes ~ [\mathbf{l}_j]_\times \enspace .
\end{equation}

\noindent The line measurement matrix $\bar{\mathsf{M}}$ is thus $3m \times 18$\footnote{
	Note that only two of the three rows of $\bar{\mathsf{M}}$ defined by Eq.~(\ref{eq:linemeasurementmatrix}) are needed for each line-line correspondence, because they are linearly dependent.
    $\bar{\mathsf{M}}$ would be only $2m \times 18$ in this case.
}. See \ref{subsec:appendix-corresp-line-line} for a derivation of Eq.~(\ref{eq:linemeasurementmatrix}).

\paragraph{Extraction of pose parameters}
First, the scale of $\bar{\mathsf{P}}'$ is corrected using Algorithm~\ref{alg:scale}, yielding $s \bar{\mathsf{P}}'$.
Then, the camera pose parameters are extracted from the right $3 \times 3$ submatrix of $s \bar{\mathsf{P}}'$, which is an estimate of a skew-symmetric matrix premultiplied by a rotation matrix (i.\,e.\ $\mathsf{R} [-\mathbf{T}]_\times$, see Eq.~(\ref{eq:lineprojectionmatrix})).
Since this is the structure of the essential matrix \citep{longuet1981computer}, we propose the algorithm of  \cite{tsai1984essential} to decompose it, as outlined in Algorithm~\ref{alg:essential-decompose}.
This completes the extraction of pose parameters.

\begin{algorithm}
	\normalsize
	\caption{Extraction of pose parameters from the estimate $\bar{\mathsf{P}}'$ of a line projection matrix (inspired by \citeauthor{tsai1984essential}, \citeyear{tsai1984essential}).}
	\label{alg:essential-decompose}
	\begin{algorithmic}[1]
    	\REQUIRE An estimate $\bar{\mathsf{P}}'$ of a line projection matrix $\bar{\mathsf{P}}$.
        \REQUIRE Corrective scale factor $s$.
        \STATE $\bar{\mathsf{P}}'_2 \leftarrow$ right $3 \times 3$ submatrix of $\bar{\mathsf{P}}'$
		\STATE $\mathsf{U} \mathsf{\Sigma} \mathsf{V}^\top \leftarrow \mathrm{SVD}(s \bar{\mathsf{P}}'_2)$
        \vspace{.5em}
		\STATE  $\mathsf{Z} \leftarrow \left[ \begin{array}{rrr}
							0 & 1 & 0 \\
							-1 & 0 & 0 \\
							0 & 0 & 0 
						\end{array} \right]$ , \quad
						$\mathsf{W} \leftarrow \left[ \begin{array}{rrr}
							0 & -1 & 0 \\
							1 & 0 & 0 \\
							0 & 0 & 1 
						\end{array} \right]$ , 
        \vspace{.75em}
		\STATEx $q \leftarrow (\mathsf{\Sigma}_{1,1} + \mathsf{\Sigma}_{2,2}) / 2$
		\STATE Compute 2 candidate solutions (A, B):
		\STATEx $\mathsf{R}_\mathrm{A} \leftarrow \mathsf{UW} \hspace{.65em} \mathrm{diag}(1 \; 1 \; \pm 1) \mathsf{V}^\top$, \quad $[\mathbf{T}]_{\times\mathrm{A}} \leftarrow q \mathsf{VZ} \hspace{.65em} \mathsf{V}^\top$
		\STATEx $\mathsf{R}_\mathrm{B} \leftarrow \mathsf{UW}^\top \mathrm{diag}(1 \; 1 \; \pm 1) \mathsf{V}^\top$, \quad $[\mathbf{T}]_{\times\mathrm{B}} \leftarrow q \mathsf{VZ}^\top \mathsf{V}^\top$
		\STATE Accept the physically plausible solution, so that the scene lies in front of the camera.
        \STATEx	$\mathsf{R} \leftarrow \mathsf{R}_\mathrm{A}$ , \quad $\mathbf{T} \leftarrow \mathbf{T}_\mathrm{A}$ \enspace or
        \STATEx $\mathsf{R} \leftarrow \mathsf{R}_\mathrm{B}$ , \quad $\mathbf{T} \leftarrow \mathbf{T}_\mathrm{B}$ \enspace .
        \ENSURE $\mathsf{R}$, $\mathbf{T}$.
    \end{algorithmic}
\end{algorithm}

The variable $q = (\mathrm{\Sigma}_{1,1} + \mathrm{\Sigma}_{2,2}) / 2$ in Algorithm~\ref{alg:essential-decompose} is an average of the first two singular values of $s\bar{\mathsf{P}}'_2$ to approximate the singular values of a properly constrained essential matrix, which should be $(q, q, 0)$.
The $\pm 1$ term in Step 4 of Algorithm~\ref{alg:essential-decompose} denotes either $+1$ or $-1$ which has to be put on the diagonal so that $\det (\mathsf{R}_\mathrm{A}) = \det (\mathsf{R}_\mathrm{B}) = 1$.

Alternative ways of extracting the camera pose parameters from $s\bar{\mathsf{P}}'$ exist, e.\,g.\ computing the closest rotation matrix $\mathsf{R}$ to the left $3 \times 3$ submatrix of $s\bar{\mathsf{P}}'_1$ and then computing  $[\mathbf{T}]_{\times} = - \mathsf{R}^\top s\bar{\mathsf{P}}'_2$. However, our experiments showed that the alternative ways are less robust when dealing with image noise. Therefore, we have chosen the solution described in Algorithm~\ref{alg:essential-decompose}.

\subsection{Algebraic outlier rejection}
\label{subsec:AOR}

In practice, mismatches of lines (i.\,e.\ outlying correspondences) often occur, which degrades the performance of camera pose estimation.
The RANSAC algorithm is commonly used to identify and remove outliers; however, as the LPnL methods work with 5 or more line correspondences, they cannot compete with the minimal (P3L) methods when plugged into a RANSAC-like framework due to the increased number of iterations required.

For this reason, an alternative scheme called Algebraic Outlier Rejection (AOR, \citeauthor{ferraz2014very}, \citeyear{ferraz2014very}) may be used instead.
It is an iterative approach integrated directly into the pose estimation procedure (specifically, into solving Eq.~(\ref{eq:noisysystem}) in Section~\ref{subsec:DLT-general} in form of iteratively reweighted least squares).
Incorrect correspondences are identified as outlying based on the residual $\epsilon_i$ of the least squares solution in Eq.~(\ref{eq:noisysystem}). Correspondences with residuals above a predefined threshold $\epsilon_\mathrm{max}$ are assigned zero weights, which effectively removes them from processing in the next iteration, and the solution is recomputed.
This is repeated until the error of the solution stops decreasing. 

The strategy for choosing $\epsilon_\mathrm{max}$ may be arbitrary, but our experiments showed that the strategy $\epsilon_\mathrm{max} = \mathrm{Q}_{j}(\epsilon_1, \ldots, \epsilon_n)$ has a good tradeoff between robustness and the number of iterations.
$\mathrm{Q}_j(\cdot)$ denotes the $j$th quantile, where $j$ decreases following the sequence (0.9, 0.8, $\ldots$ , 0.3)
and then it remains constant 0.25 until error of the solution stops decreasing.
This strategy usually leads to approximately 10 iterations.

\begin{rmk}
	It is important \emph{not} to prenormalize the data in this case because it will impede the identification of outliers. Prenormalization of inliers should be done just before the last iteration.
\end{rmk}

Compared to RANSAC, the benefit of this approach is a low runtime independent of the fraction of outliers.
On the other hand, the break-down point is somewhere between 40\,\% and 70\,\% of outliers, depending on the underlying LPnL method, whereas RANSAC can handle any fraction of outliers in theory.

\section{DLT-Combined-Lines}
\label{sec:DLT-Combined-Lines}

In this section, we introduce the novel method DLT-Combined-Lines. It is a combination of DLT-Lines and DLT-Plücker-Lines, exploiting the redundant representation of 3D structure in the form of both 3D points and 3D lines.
The 2D structure is represented by 2D lines.
The outcome is a higher accuracy of the camera pose estimates, smaller reprojection error, and lower number of lines required.

The central idea is to merge two systems of linear equations, which share some unknowns, into one system.
The unknowns are entries of the point projection matrix $\dot{\mathsf{P}}$ and the line projection matrix $\bar{\mathsf{P}}$.
The two systems defined by Eq.~(\ref{eq:DLT-Lines}) and (\ref{eq:DLT-Plucker-Lines}) can be merged so that the set of unknowns of the resulting system is formed by the union of unknowns of both systems. It can be observed that the shared unknowns reside in the left $3 \times 3$ submatrices of $\dot{\mathsf{P}}$ and $\bar{\mathsf{P}}$. If unknowns of the resulting system are arranged in a feasible manner, a new $3 \times 7$ matrix $\ddot{\mathsf{P}}$ can be constructed, which is a ``union'' of $\dot{\mathsf{P}}$ and $\bar{\mathsf{P}}$:
\begin{equation}
  \begin{rcases*}
  	\dot{\mathsf{P}} \approx
    \left[ ~
    	\mathsf{R} \hspace{1.1em} {-}\mathsf{R}\mathbf{T}
    \hspace{1em} \right] \\
    \bar{\mathsf{P}} \approx
    \left[ ~
    	\mathsf{R} \hspace{.9em} \mathsf{R}[-\mathbf{T}]_{\times} 
    \right] \hspace{.7em}
  \end{rcases*}
  	\hspace{1em}
   \ddot{\mathsf{P}} \approx
	\left[ \begin{array}{ccccc}
		\mathsf{R} & & -\mathsf{R}\mathbf{T}  & & \mathsf{R} [-\mathbf{T}]_{\times} 
	\end{array} \right]
    \label{eq:combinedprojectionmatrix}
\end{equation}

\noindent We call the matrix a ``combined projection matrix'', because it allows us to write projection equations for point-line, line-line, and even point-point correspondences, as follows:
\begin{eqnarray}
  	\label{eq:DLT-Combined-Lines-points-lines}
	\mathbf{l}^\top \ddot{\mathsf{P}} \left( \hspace{0.8em} \mathbf{X}^\top \hspace{1em} 0 \;~ 0 \;~ 0 ~ \right)^\top = 0 \enspace , \\
  	\label{eq:DLT-Combined-Lines-lines-lines}
	\mathbf{l} \approx \ddot{\mathsf{P}} \left( ~ \mathbf{U}^\top \; 0 \hspace{1.6em} \mathbf{V}^\top \hspace{0.6em} \right)^\top \enspace , \hspace{1.6em} \\
    \label{eq:DLT-Combined-Lines-points-points}
	\mathbf{x} \approx \ddot{\mathsf{P}} \left( \hspace{0.9em} \mathbf{X}^\top \hspace{1em} 0 \;~ 0 \;~ 0 ~ \right)^\top \enspace . \hspace{1.7em}
\end{eqnarray}

\noindent These equations can then be used to estimate $\ddot{\mathsf{P}}$ linearly from the correspondences, as shown in detail in Section~\ref{subsec:DLT-combined-linear-estim}.

Higher accuracy of pose estimates using the proposed method stems from the fact that the left-most $\mathsf{R}$ in $\ddot{\mathsf{P}}$ is determined by twice as many equations, and also from the fact that $\ddot{\mathsf{P}}$ contains multiple estimates of $\mathsf{R}$ and $\mathbf{T}$.
This is further investigated in Section~\ref{subsec:DLT-combined-extraction}.

The other benefit is that the method requires only 5 lines (and 10 points across them) -- less then DLT-Plücker-Lines and even less then DLT-Lines.
To explain why, we first define the following matrices: the left-most $3 \times 3$ submatrix of $\ddot{\mathsf{P}}$ is denoted $\ddot{\mathsf{P}}_1$, the middle $3 \times 1$ submatrix (column vector) is denoted $\ddot{\mathbf{P}}_2$, and the right-most $3 \times 3$ submatrix is denoted $\ddot{\mathsf{P}}_3$.
\begin{equation}
  \ddot{\mathsf{P}} = 
    \left[ \begin{array}{ccccc}
		\mathsf{R} & & -\mathsf{R}\mathbf{T}  & & \mathsf{R} [-\mathbf{T}]_{\times} 
	\end{array} \right] =
    \left[ \begin{array}{ccccc}
		\ddot{\mathsf{P}}_1 & & \ddot{\mathbf{P}}_2  & &\ddot{\mathsf{P}}_3 
	\end{array} \right]
  \label{eq:combinedprojmatrix-submatrices}
\end{equation}

\noindent $\ddot{\mathsf{P}}$ has 21 entries, but since it encodes the camera pose, it has only 6 DoF.
This means it has 14 nonlinear constraints (homogeneity of the matrix accounts for the 1 remaining DoF).
Ignoring the nonlinear constraints, which are not taken into account during the least squares estimation, $\ddot{\mathsf{P}}$ has 20 DoF.
Each point-line correspondence generates 1 independent linear equation (\ref{eq:DLT-Combined-Lines-points-lines}) and each line-line correspondence generates 2 independent linear equations (\ref{eq:DLT-Combined-Lines-lines-lines}).
Since $\ddot{\mathbf{P}}_2$ is determined only by point-line correspondences and since it has 3 DoF, at least 3 3D points are required to fully determine it.
An analogy holds for $\ddot{\mathsf{P}}_3$: since it is determined only by line-line correspondences and since it has 9 DoF, at least 5 (in theory 4\sfrac{1}{2}) 3D lines are required to fully determine it.
The required number of $m$ line-line correspondences and $n$ point-line correspondences is thus $m \!=\! 9$, $n \!=\! 3$, or $m \!=\! 5$, $n \!=\! 10$, or something in between satisfying the inequality $(n+2m) \ge 20$.
In such minimal cases, the points must be distributed equally among the lines, i.\,e.\ each point or two must lie on a different line; otherwise, the system of equations would be underdetermined.

We now proceed with the description of the algorithm.
Please notice that the prenormalization procedure will be described in Section~\ref{subsec:DLT-combined-prenormalization}, i.\,e.\ \emph{after} the definition of a measurement matrix in Section~\ref{subsec:DLT-combined-linear-estim}, because prenormalization is strongly motivated by its structure.

\subsection{Linear estimation of the combined projection matrix}
\label{subsec:DLT-combined-linear-estim}

The combined projection matrix $\ddot{\mathsf{P}}$ and its estimate $\ddot{\mathsf{P}}'$ are $3 \times 7$, so the combined measurement matrix $\ddot{\mathsf{M}}$ has 21 columns.
The number of its rows depends on the number of $n$ point-line correspondences $\mathbf{X}_i \leftrightarrow \mathbf{l}_i$, $(i = 1 \dots n)$, and on the number of $m$ line-line correspondences $\mathbf{L}_j \leftrightarrow \mathbf{l}_j$, $(j = n+1 \dots n+m)$.
The minimal values of $n$ and $m$ depend on each other and must satisfy the inequality $(n+2m) \ge 20$.
Each point-line correspondence (\ref{eq:DLT-Combined-Lines-points-lines}) leads to one row of $\ddot{\mathsf{M}}$, and each line-line correspondence (\ref{eq:DLT-Combined-Lines-lines-lines}) gives rise to three rows of $\ddot{\mathsf{M}}$ (Matlab notation is used to index the matrix elements):
\begin{eqnarray}
    \label{eq:combinedmeasurementmatrix-points}
    \ddot{\mathsf{M}}_{\scriptscriptstyle (i,~:)} \hspace{4.8em} &=&
    	(\mathbf{X}_i^\top \;~ 0 \;~ 0 \;~ 0) ~ \otimes ~ \mathbf{l}_i^\top \enspace , \\
    \label{eq:combinedmeasurementmatrix-lines}
	\ddot{\mathsf{M}}_{\scriptscriptstyle (3j-n-2 ~ : ~ 3j-n,~:)} &=&
    	(\mathbf{U}_j^\top \;~ 0 \;~ \mathbf{V}_j^\top) ~ \otimes ~ [\mathbf{l}_j]_\times \enspace .
\end{eqnarray}

\noindent The combined measurement matrix $\ddot{\mathsf{M}}$ is thus $(n+3m) \times 21$\footnote{
	Note that only two of the three rows of $\ddot{\mathsf{M}}$ defined by Eq.~(\ref{eq:combinedmeasurementmatrix-lines}) are needed for each line-line correspondence, because they are linearly dependent.
    Our experiments showed that using all three rows brings no advantage, so we use only two of them in practice.
    In this case, $\ddot{\mathsf{M}}$ is only $(n+2m) \times 21$.
}.
See \ref{sec:appendix-corresp} for derivations of Eq.~(\ref{eq:combinedmeasurementmatrix-points}) and (\ref{eq:combinedmeasurementmatrix-lines}).
The combined measurement matrix $\ddot{\mathsf{M}}$ can also be constructed by stacking and aligning $\dot{\mathsf{M}}$ and $\bar{\mathsf{M}}$: 
\begin{equation}
	\label{eq:combinedmeasurementmatrix-stacked}
	\ddot{\mathsf{M}} =
    	{\arraycolsep=.5em\def\arraystretch{1.3}
    	\left[ \begin{array}{ccl}
			\multicolumn{2}{c}{\dot{\mathsf{M}}_{\scriptscriptstyle n \times 12}} & \mathsf{0}_{\scriptscriptstyle n \times 9} \\

			\bar{\mathsf{M}}_{\scriptscriptstyle (:,1:9)} & \mathsf{0}_{\scriptscriptstyle 3m \times 3} & \bar{\mathsf{M}}_{\scriptscriptstyle (:, 10:18)}
		\end{array} \right]
        }
\end{equation}

\begin{rmk}
  \label{rmk:mmatrix-scale}
  It is advisable to scale both $\dot{\mathsf{M}}$ and $\bar{\mathsf{M}}$ so that the sums of squares of their entries are equal.
  (If they were not, it would negatively affect the scales of those parts of the solution $\ddot{\mathbf{p}} = \mathrm{vec}(\ddot{\mathsf{P}})$, which are determined exclusively by either $\dot{\mathsf{M}}$ or $\bar{\mathsf{M}}$, but not by both of them. These are the entries 10-12 and 13-21 of $\ddot{\mathbf{p}}$, which contain estimates of translation. See the middle and right part of $\ddot{\mathsf{P}}$ in Eq.~(\ref{eq:combinedprojmatrix-submatrices}).)
\end{rmk}

\begin{rmk}
  The method can easily be extended to point-point correspondences (\ref{eq:DLT-Combined-Lines-points-points}) by adding extra rows to $\ddot{\mathsf{M}}$. Each of the $p$ point-point correspondences $\mathbf{X}_k \leftrightarrow \mathbf{x}_k$, $(k = n+m+1 \dots n+m+p)$ generates three rows
  \begin{eqnarray}
      \label{eq:combinedmeasurementmatrix-pointspoints}
      \ddot{\mathsf{M}}_{\scriptscriptstyle (3k-n-m-2 ~ : ~ 3k-n-m,~:)} &=& (\mathbf{X}_i^\top \;~ 0 \;~ 0 \;~ 0) ~ \otimes ~ [\mathbf{x}_i]_\times \enspace ,
  \end{eqnarray}
  \noindent two of which are linearly independent.
  See \ref{subsec:appendix-corresp-point-point} for a derivation of Eq.~(\ref{eq:combinedmeasurementmatrix-pointspoints}).
\end{rmk}

\subsection{Prenormalization}
\label{subsec:DLT-combined-prenormalization}

Prenormalization of 2D lines is rather complicated.
The problem is that a 2D line $\mathbf{l}$ is in the direct form and on the opposite side from the line projection matrix $\bar{\mathsf{P}}$ in Eq.~(\ref{eq:DLT-Combined-Lines-lines-lines}), and it is in the transposed form and on the same side as the point projection matrix $\dot{\mathsf{P}}$ in Eq.~(\ref{eq:DLT-Combined-Lines-points-lines}).
Thus, when undoing the effect of a prenormalizing 2D transformation $\mathsf{t}$, the inverse transformation is $\mathsf{t}^{-1}$ for $\bar{\mathsf{P}}$, and $\mathsf{t}^\top$ for $\dot{\mathsf{P}}$.
Since both $\dot{\mathsf{P}}$ and $\bar{\mathsf{P}}$ are parts of $\ddot{\mathsf{P}}$, both inverse transformations must be identical ($\mathsf{t}^\top = \mathsf{t}^{-1}$).
However, this only holds for a 2D rotation, which is practically useless as a prenormalizing transformation.
We thus suggest not prenormalizing 2D lines at all.

Prenormalization of 3D points and 3D lines is also nontrivial, because transformations of 3D space affect the coordinates of points and lines differently. However, it can be achieved by pursuing the goal from the beginning of Section~\ref{subsec:DLT-general}: to center the data around the origin by translation, and to scale them so that an average coordinate has the absolute value of 1.

Please note that translation and scaling affects only the $\mathbf{U}$ part of a 3D line $\mathbf{L}$, and only the $(X_1 ~ X_2 ~ X_3)^\top$ part of a 3D point $\mathbf{X}$.
Therefore, (i) the unaffected parts ($\mathbf{V}$ and $X_4$) must be adjusted beforehand: Each 3D line and each 3D point is normalized by multiplication by a non-zero scale factor, so that $||\mathbf{V}|| = \sqrt{3}$, and $X_4 = 1$. Note that this adjustment does \emph{not} change the spatial properties of 3D points/lines.
Then, (ii) translation is applied to center the 3D points around the origin\footnote{
	Another possible translation is to center the 3D lines using the Generalized Weiszfeld algorithm \citep{aftab1lqclosest}.
    However, our experiments showed that the two possible translations yield nearly identical robustness of the method.
   	We thus suggest to translate the 3D structure to the centroid of points, because its computation is cheaper.
}.
Although the translation is intuitively correct (it results in zero mean of 3D points), it is not optimal in terms of entries of the measurement matrix (joint zero mean of $(X_1 ~ X_2 ~ X_3)^\top$ and $\mathbf{U}$).
Therefore, (iii) another translation is applied to minimize the average magnitude of $(X_1 ~ X_2 ~ X_3)^\top$ and $\mathbf{U}$.
Finally, (iv) anisotropic scaling is applied so that the average magnitudes of all $X_1$ and $L_1$, $X_2$ and $L_2$, $X_3$ and $L_3$, $X_4$ and $\mathbf{V}$ are equal, i.\,e.\ $\overline{|X_1|} + \overline{|L_1|} = \overline{|X_2|} + \overline{|L_2|} = \overline{|X_3|} + \overline{|L_3|} = \overline{|X_4|} + \overline{(|L_4|, |L_5|, |L_6|)}$.
This also ensures that the corresponding blocks of the measurement matrix $\ddot{\mathsf{M}}$ will have equal average magnitude.
The very last step of prenormalization (v) is not applied to the input primitives, but to the measurement matrix after its construction. Its point- and line-related parts $\dot{\mathsf{M}}$ and $\bar{\mathsf{M}}$ should be scaled as stated in Remark~\ref{rmk:mmatrix-scale} above.

The effects of individual stages of prenormalization on accuracy of the proposed method are experimentally evaluated in Section~\ref{subsec:results-prenormalization}.

\subsection{Extraction of pose parameters}
\label{subsec:DLT-combined-extraction}

First, the scale of $\ddot{\mathsf{P}}'$ is corrected using Algorithm~\ref{alg:scale}, yielding $s \ddot{\mathsf{P}}'$.
The estimates of $\mathsf{R}$ and $\mathbf{T}$ are doubled in $s\ddot{\mathsf{P}}'$, which can be exploited to estimate the camera pose more robustly.
In the following, we use the definitions of submatrices $\ddot{\mathsf{P}}_1$, $\ddot{\mathbf{P}}_2$, and $\ddot{\mathsf{P}}_3$ from Eq.~(\ref{eq:combinedprojmatrix-submatrices}).
The first estimate of $\mathsf{R}$ is in the direct form in $s\ddot{\mathsf{P}}'_1$, from which it can be extracted using Algorithm~\ref{alg:orthogonalization}, yielding $\mathsf{R}_1$.
The first estimate of $\mathbf{T}$ is in $s\ddot{\mathbf{P}}'_2$, premultiplied by $-\mathsf{R}$. It can be recovered as $\mathbf{T}_2 = -\mathsf{R}^\top_1 s \ddot{\mathbf{P}}'_2$.
The second estimates of $\mathsf{R}$ and $\mathbf{T}$ are in the form of an essential matrix in $s\ddot{\mathsf{P}}'_3$, from which they can be extracted using Algorithm~\ref{alg:essential-decompose}, yielding $\mathsf{R}_3$ and $\mathbf{T}_3$.

Now, the question is how to combine $\mathsf{R}_1$, $\mathsf{R}_3$, and $\mathbf{T}_2$, $\mathbf{T}_3$.
Our experiments showed that $\mathsf{R}_1$ is usually more accurate than $\mathsf{R}_3$, probably because it is determined by twice as many equations (generated by both line-line and point-line correspondences).
The experiments also showed that $\mathbf{T}_2$ is usually more accurate than $\mathbf{T}_3$.
We hypothesize this is because $\ddot{\mathbf{P}}'_2$ has no redundant DoF, contrary to  $\ddot{\mathsf{P}}'_3$, which has 3 redundant DoF.
However, the estimates can be combined so that the result is even more accurate. Since the error vectors of $\mathbf{T}_2$ and $\mathbf{T}_3$ tend to have opposite direction, a suitable interpolation between them can produce a more accurate position estimate
\begin{equation}
	\mathbf{T} = k \cdot \mathbf{T}_2 + (1-k) \cdot \mathbf{T}_3 \enspace .
\end{equation}
  
\noindent The value of $k$ should be between 0 and 1. Based on grid search, an optimal value of 0.7 has been found (the error function has a parabolic shape).
Regarding the rotation estimates, the grid search discovered $\mathsf{R}_1$ is indeed more accurate than $\mathsf{R}_3$. However, $\mathsf{R}_1$ is not fully ``compatible'' with $\mathbf{T}$ in terms of reprojection error\footnote{As an example, imagine a camera located left to its ground truth position \emph{and} oriented even more left.}. Interpolating between $\mathsf{R}_1$ and $\mathsf{R}_3$ yields an orientation $\mathsf{R}$ ``compatible'' with $\mathbf{T}$:
\begin{equation}
	\mathsf{R} = \mathsf{R}_1 \cdot \exp (k \cdot \log(\mathsf{R}^\top_1 \mathsf{R}_3)) \enspace .
\end{equation}
  
\noindent Here, ``exp'' and ``log'' denote matrix exponential and matrix logarithm, respectively.
The whole pose extraction procedure is summarized in Algorithm~\ref{alg:combined-extract}.

\begin{algorithm}
	\normalsize
	\caption{Extraction of pose parameters from the estimate $\ddot{\mathsf{P}}'$ of a combined projection matrix.}
	\label{alg:combined-extract}
	\begin{algorithmic}[1]
    	\REQUIRE An estimate $\ddot{\mathsf{P}}'$ of a line projection matrix $\ddot{\mathsf{P}}$.
        \REQUIRE Corrective scale factor $s$.
        \STATE $\left[ \begin{array}{ccccc}
					\ddot{\mathsf{P}}'_1 & & \ddot{\mathbf{P}}'_2  & &\ddot{\mathsf{P}}'_3 
				\end{array} \right] \leftarrow \ddot{\mathsf{P}}'$ // divide into submatrices
        \STATE Extract $\mathsf{R}_1$ from $\ddot{\mathsf{P}}'_1$ using Algorithm~\ref{alg:orthogonalization}.
        \STATE $\mathbf{T}_2 = -\mathsf{R}^\top_1 s \ddot{\mathbf{P}}'_2$
        \STATE Extract $\mathsf{R}_3$, $\mathbf{T}_3$ from $\ddot{\mathsf{P}}'_3$ using Algorithm~\ref{alg:essential-decompose}.
        \STATE $\mathsf{R} = \mathsf{R}_1 \cdot \exp (k \cdot \log(\mathsf{R}^\top_1 \mathsf{R}_3))$
        \STATEx $\mathbf{T} = k \cdot \mathbf{T}_2 + (1-k) \cdot \mathbf{T}_3$
        \ENSURE $\mathsf{R}$, $\mathbf{T}$.
    \end{algorithmic}
\end{algorithm}

\section{Experimental results}
\label{sec:results}

The accuracy of pose estimates, the computational efficiency of the methods, and their robustness to image noise and to outliers were measured.
Accuracy of pose estimates was expressed in terms of position error and orientation error of the camera, and in terms of reprojection error of the lines, as each error measure suits different applications.
For example, robot localization requires a minimal position error, visual servoing requires both position and orientation error to be small, while augmented reality applications prioritize a small reprojection error.

The proposed algorithm was evaluated and compared with the following state-of-the-art methods:

\begin{enumerate}
	\item \textbf{Ansar}, the method by \cite{ansar2003linear}, implementation from \cite{xu2016pnl}, results shown in black \markAnsar\,.
	\item \textbf{Mirzaei}, the method by \cite{mirzaei2011globally}, results shown in red \markMirzaei\,.
	\item \textbf{RPnL}, the method by \cite{zhang2012rpnl}, results shown in dark blue \markRPnL\,.
    \item \textbf{ASPnL}, the method by \cite{xu2016pnl}, results shown in light blue \markASPnL\,.
    \item \textbf{LPnL\_Bar\_LS}, the method by \cite{xu2016pnl}, results shown in teal \markLPnLBarLS\,.
    \item \textbf{LPnL\_Bar\_ENull}, the method by \cite{xu2016pnl}, results shown in blue-green \markLPnLBarENull\,.
    \item \textbf{DLT-Lines}, the method by \citet[\DLTlinesPage]{hartley2004multiple}, our implementation, results shown in purple \markDLTlines\,.
	\item \textbf{DLT-Plücker-Lines}, the method by \cite{Pribyl2015}, our implementation, results shown in green \markDLTplucker\,.
    \item \textbf{DLT-Combined-Lines}, the proposed method, results shown in orange \markDLTcombined\,.
\end{enumerate}

\noindent All methods were implemented in Matlab. The implementations originate from the respective authors, if not stated otherwise.

First, we evaluate accuracy, robustness, and efficiency of the methods using synthetic lines. Then, we evaluate accuracy of pose estimates using rendered images and real data.

\subsection{Synthetic lines}
\label{subsec:results-synthetic}

Monte Carlo simulations with synthetic lines were performed under the following setup: at each trial, $m$ 3D line segments were generated by randomly placing $n = 2m$ line endpoints inside a cube spanning $10^3$\,m which was centered at the origin of the world coordinate system. For the methods which work with 3D points, the endpoints were used.
A virtual pinhole camera with image size of $640 \times 480$ pixels and focal length of 800\,pixels was placed randomly at the distance of 25\,m from the origin.
The camera was then oriented so that it looked directly at the origin, having all 3D line segments in its field of view.
The 3D line segments were projected onto the image plane. Coordinates of the 2D endpoints were then perturbed with independent and identically distributed Gaussian noise with standard deviation of $\sigma$ pixels.
1000 trials were carried out for each combination of $m$, $\sigma$ parameters.

Accuracy of pose estimation and robustness to image noise of each method was evaluated by measuring the estimated and true camera pose while varying $m$ and $\sigma$ similarly to~\cite{mirzaei2011globally}.
The \emph{position error} $\mathrm{\Delta\Tau} = ||\mathbf{T}' - \mathbf{T}||$ is the distance from the estimated position $\mathbf{T}'$ to the true position $\mathbf{T}$.
The \emph{orientation error} $\mathrm{\Delta\Theta}$  was calculated as follows. The difference between the true and estimated rotation matrix ($\mathsf{R}^\top \mathsf{R}'$) is converted to axis-angle representation ($\mathbf{E}$, $\mathrm{\Theta}$) and the absolute value of the difference angle $|\mathrm{\Theta}|$ is considered as the orientation error.
The \emph{reprojection error} $\mathrm{\Delta\pi}$ is an integral of squared distance between points on the image line segment and the projection of an infinite 3D line according to \cite{taylor1995structure}, averaged\footnote{
	Please note that \cite{taylor1995structure} defined the reprojection error as a \emph{sum} over all individual lines, which makes it dependent on the number of lines.
} over all individual lines.

\afterpage{%
	\clearpage
	\begin{landscape}
		\begin{figure*}
        	\vspace{-3.85cm}\hspace{-3.8cm}
			\includegraphics[width=1.4\linewidth]{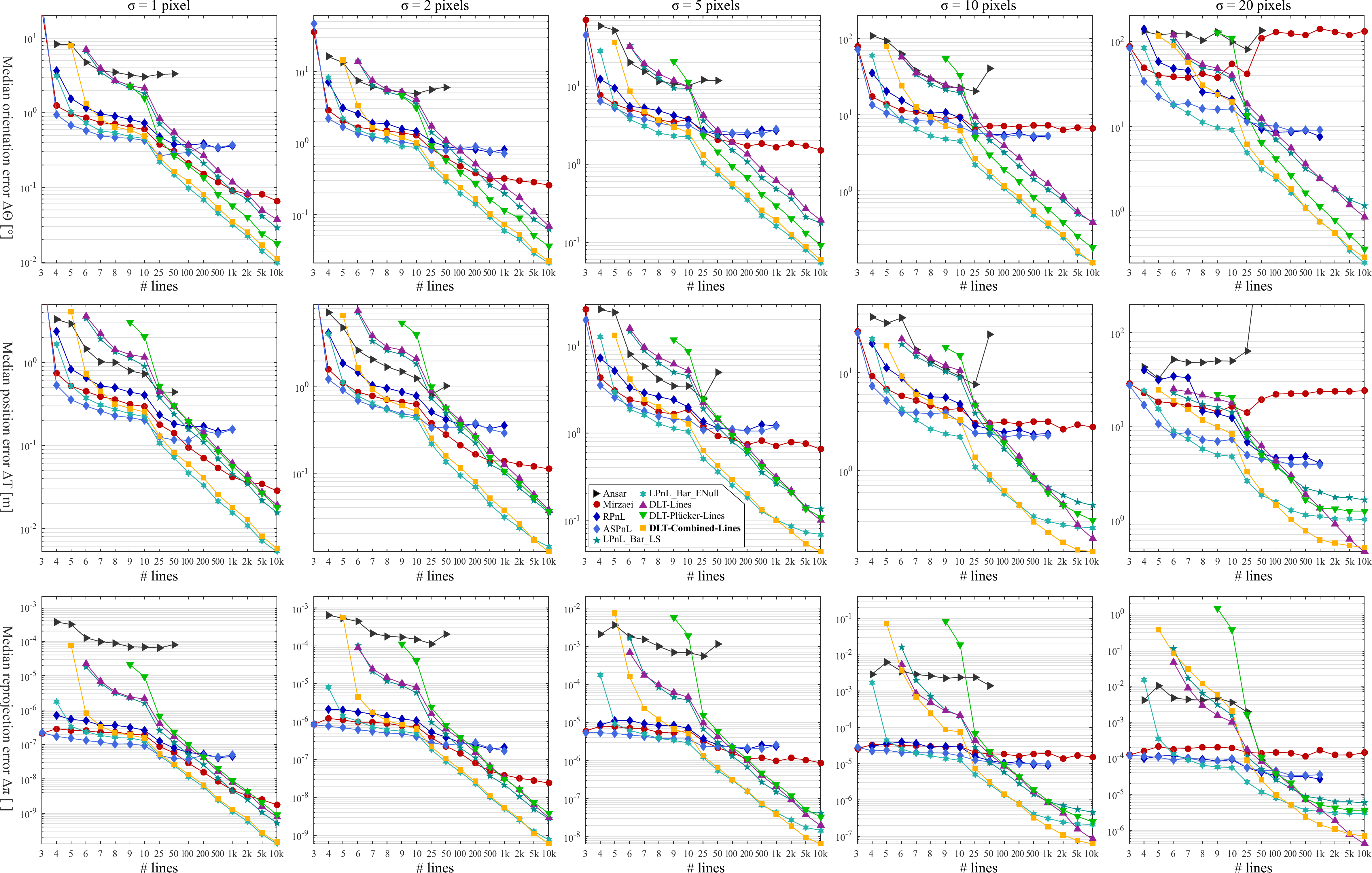}
		    \caption{
				Experiments with synthetic lines. Median errors in estimated camera pose as a function of the number of lines, computed from 1000 trials. Orientation errors ($\mathrm{\Delta\Theta}$, \emph{top row}), position errors ($\mathrm{\Delta\Tau}$, \emph{middle row}) and reprojection errors ($\mathrm{\Delta\pi}$, \emph{bottom row}) are depicted for various levels of image noise ($\sigma$ = 1\,px -- 20\,px, \emph{from left to right}). All vertical axes are logarithmic.
			}
			\label{fig:errors}
		\end{figure*}
	\end{landscape}
	\clearpage
}

The results showing accuracy of the methods and their robustness to image noise are depicted in Fig.~\ref{fig:errors}.
Errors for each method are plotted from the minimal number of lines to 10,000 lines (or less, if the method runs too long or if it has impractical memory requirements).
In the following text, the method names are often augmented with their plot marks to ease referencing into result charts.

Our results show high sensitivity to noise of \textbf{Ansar}\,\markAnsar. Even under slight image noise $\sigma$ = 1\,px, the measured accuracy is poor.
The other non-LPnL methods (\textbf{Mirzaei}\,\markMirzaei, \textbf{RPnL}\,\markRPnL, \textbf{ASPnL}\,\markASPnL) outperform the LPnL methods for low number of lines (3 -- 10), as expected. \textbf{ASPnL} is the most accurate among them.
An exception is the LPnL method \textbf{LPnL\-\_Bar\-\_ENull}\,\markLPnLBarENull, the accuracy of which is close to \textbf{ASPnL}\,\markASPnL. It even outperforms \textbf{ASPnL} in the case of medium and strong image noise ($\sigma$ = 5 -- 20\,px), see Fig.~\ref{fig:errors}.

For high number of lines (100 -- 10,000), the LPnL methods outperform the non-LPnL ones.
\textbf{LPnL\-\_Bar\-\_ENull}\,\markLPnLBarENull\, and \textbf{DLT-Combined-Lines}\,\markDLTcombined\, are significantly most accurate in both orientation and position estimation, and they also yield the lowest reprojection error.
With increasing number of lines, accuracy of the LPnL methods further increases, while the errors of the non-LPnL methods (\textbf{Mirzaei}\,\markMirzaei, \textbf{RPnL}\,\markRPnL, \textbf{ASPnL}\,\markASPnL) do not fall below a certain level. This gets more obvious with increasing levels of noise, see Fig.~\ref{fig:errors}.
Each of the LPnL methods also eventually reaches its limit, as it can be seen in the bottom right area of Fig.~\ref{fig:errors}.
However, the accuracy limits of non-LPnL methods lag behind the limits of LPnL methods.

\textbf{DLT-Lines}\,\markDLTlines\, and \textbf{LPnL\_Bar\_LS}\,\markLPnLBarLS\, behave nearly identically, the latter being slightly more accurate.
The only difference between the two is the latter's use of barycentric coordinates, to which the slight improvement in results can probably be attributed.
However, \textbf{DLT-Lines} proves to be more accurate in position estimation and reprojection under strong image noise.
\textbf{DLT-Plücker-Lines}\,\markDLTplucker\, performs comparably with the two aforementioned methods
for 25 or more lines.

The best accuracy on many lines is achieved by the \textbf{LPnL\-\_Bar\-\_ENull}\,\markLPnLBarENull\, and \textbf{DLT-Combined-Lines}\,\markDLTcombined\, methods, being the best in all criteria.
While they are comparable in orientation estimation, \textbf{DLT-Combined-Lines} outperforms \textbf{LPnL\-\_Bar\-\_ENull} in estimation of camera position and in reprojection for many lines. The higher accuracy of \textbf{DLT-Combined-Lines} is most apparent under strong image noise, see the right part of Fig.~\ref{fig:errors}.

The distributions of errors of the individual methods over all 1000 trials are provided in the supplementary material.

\subsubsection{Quasi-singular line configurations}
\label{subsubsec:results-singular}

Methods for pose estimation are known to be prone to singular or quasi-singular configurations of 3D primitives.
Therefore, the robustness of the methods to quasi-singular line configurations was also evaluated. The setup from Section~\ref{subsec:results-synthetic} was used with the number of lines fixed to $m = 200$, and standard deviation of image noise fixed to $\sigma$ = 2\,px.

\paragraph{Limited number of line directions}
Lines were generated in three different scenarios: 2 random directions, 3 random directions, and 3 orthogonal directions. The methods of \textbf{Ansar} and \textbf{Mirzaei} do not work in either case. \textbf{RPnL} and \textbf{ASPnL} do work, but are susceptible to failure. \textbf{DLT-Plücker-Lines} and \textbf{DLT-Combined-Lines} do not work in the case of 2 directions, work unreliably in the case of 3 directions, and begin to work flawlessly if the 3 directions are mutually orthogonal. \textbf{DLT-Lines}, \textbf{LPnL\-\_Bar\-\_LS} and \textbf{LPnL\-\_Bar\-\_ENull} work well in all cases.

\paragraph{Near-planar line distribution}
Lines were generated inside a cube spanning $10^3$\,m, but the cube was progressively flattened until it became a plane. Nearly all methods start to degrade their accuracy when flatness of the ``cube'' reaches a ratio of 1:10 and perform noticeably worse at the ratio of 1:100. \textbf{Mirzaei}, all three \textbf{DLT-based} methods and \textbf{LPnL\-\_Bar\-\_LS} mostly stop working. \textbf{RPnL} and \textbf{ASPnL} do work, but often fail. The only fully working method is \textbf{LPnL\-\_Bar\-\_ENull}.

\paragraph{Near-concurrent line distribution}
Lines were generated randomly, but an increasing number of lines were forced to intersect at a random point inside the cube until all lines were concurrent. \textbf{Mirzaei}, \textbf{RPnL}, \textbf{ASPnL} and \textbf{LPnL\-\_Bar\-\_LS} degrade their accuracy progressively, although \textbf{ASPnL} and \textbf{LPnL\-\_Bar\-\_LS} are reasonably accurate even in the fully concurrent case. The \textbf{DLT-based} methods work without any degradation as long as 3 or more lines are non-concurrent. \textbf{LPnL\-\_Bar\-\_ENull} also works without degradation in the fully concurrent case.

To sum up, behavior of the proposed method, \textbf{DLT-Combined-Lines}, is inherited from its two predecessor methods \textbf{DLT-Lines} and \textbf{DLT-Plücker-Lines}. Their accuracy is degraded:
\begin{itemize}[nosep]
  \item If the lines tend to be \emph{planar} (flatness $\approx$ 1:10 or more).
  \item If there are fewer than 3 \emph{non-concurrent} lines.
  \item If the lines are organized into \emph{3 or less directions} (DLT-Lines works in this case, but DLT-Plücker-Lines and DLT-Combined-Lines work only if the 3 directions are orthogonal).
\end{itemize}
For the sake of brevity, charts depicting errors in the quasi-singular cases are part of the supplementary material.

\subsection{Real data}
\label{subsec:results-real}

\begin{figure*}
	\centering
	\makebox[\textwidth][c]{\includegraphics[width=1.4\linewidth]{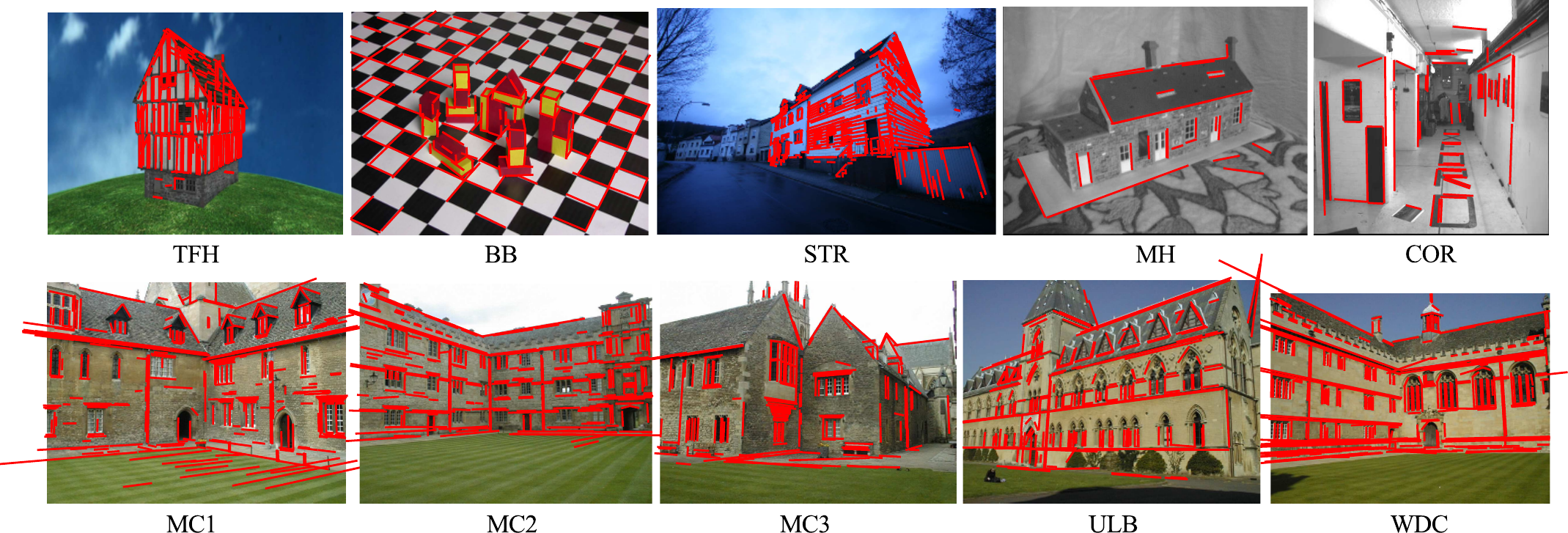}}
    \caption{
		Example images from used datasets. The images are overlaid with reprojections of 3D line segments using the camera pose estimated by the proposed method DLT-Combined-Lines.
	}
	\label{fig:real-images}
\end{figure*}

Ten datasets were utilized, which contain images with detected 2D line segments, reconstructed 3D line segments, and camera projection matrices.
Example images from the datasets are shown in Fig.~\ref{fig:real-images}, and their characteristics are summarized in Table~\ref{tab:real}.
The Timberframe House dataset contains rendered images, while the rest contain real images captured by a physical camera.  
The Building Blocks and Model House datasets capture small-scale objects on a table, the Corridor dataset captures an indoor corridor, and the other six datasets capture exteriors of various buildings.
The Building Blocks dataset is the most challenging because many line segments lie on the common plane of a chessboard.

\begin{table}
	\caption{Image sequences used in the experiments with real data.}
    \label{tab:real}
    \centering
	\begin{tabular}{lrclrr}
      \hline\noalign{\smallskip}
      Sequence & Source & & Abrv. & \#imgs. & \#lines\\
      \noalign{\smallskip}
      \hline
      \noalign{\smallskip}
      Timberframe House  & MPI\textsuperscript{\textdagger} & & TFH & 72 &  828\\
      Building Blocks    & MPI\textsuperscript{\textdagger} & & BB  & 66 &  870\\
      Street             & MPI\textsuperscript{\textdagger} & & STR & 20 & 1841\\
      Model House        & VGG\textsuperscript{\textdaggerdbl} & & MH  & 10 &   30\\
      Corridor           & VGG\textsuperscript{\textdaggerdbl} & & COR & 11 &   69\\
      Merton College I   & VGG\textsuperscript{\textdaggerdbl} & & MC1 &  3 &  295\\
      Merton College II  & VGG\textsuperscript{\textdaggerdbl} & & MC2 &  3 &  302\\
      Merton College III & VGG\textsuperscript{\textdaggerdbl} & & MC3 &  3 &  177\\
      University Library & VGG\textsuperscript{\textdaggerdbl} & & ULB &  3 &  253\\
      Wadham College     & VGG\textsuperscript{\textdaggerdbl} & & WDC &  5 &  380\\
      \hline
      \multicolumn{6}{l}{\textsuperscript{\textdagger}\footnotesize{MPI dataset \url{http://resources.mpi-inf.mpg.de/LineReconstruction/}}}\\
      \multicolumn{6}{l}{\textsuperscript{\textdaggerdbl}\footnotesize{VGG dataset \url{http://www.robots.ox.ac.uk/\textasciitilde vgg/data/data-mview.html}}}
	\end{tabular}
\end{table}

\begin{figure*}
	\centering
		\makebox[\textwidth][c]{\includegraphics[width=1.4\linewidth]{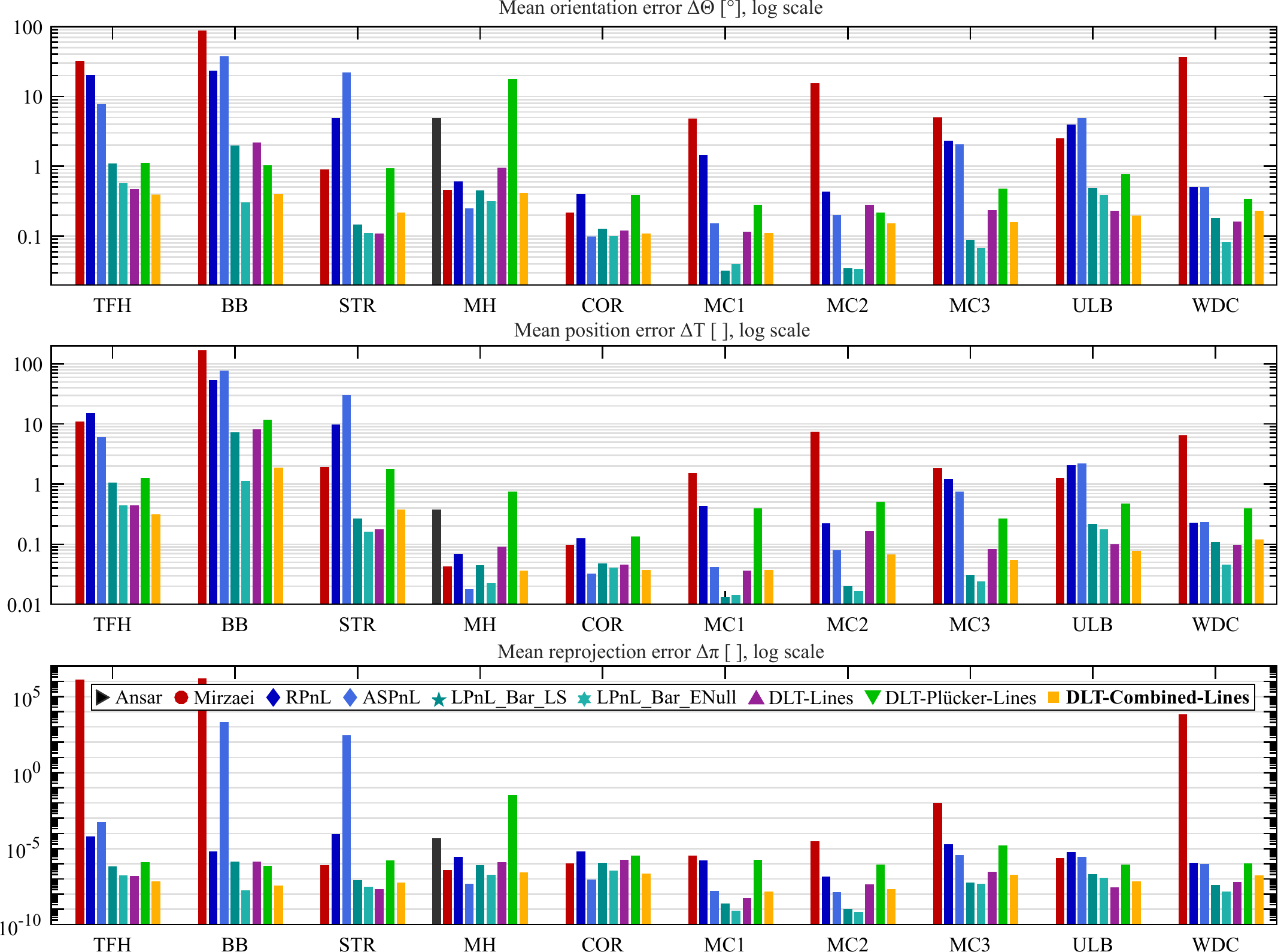}}
    \caption{Experiments with real data. Mean orientation errors ($\mathrm{\Delta\Theta}$,\emph{top}), position errors ($\mathrm{\Delta\Tau}$, \emph{middle}) and reprojection errors ($\mathrm{\Delta\pi}$, \emph{bottom}) on individual image sequences. All vertical axes are logarithmic.}
	\label{fig:real-charts}
\end{figure*}

Each PnL method was run on the data, and the errors in camera orientation, camera position and reprojection of lines were averaged over all images in each sequence.
The mean errors achieved by all methods on individual datasets are given in Table~\ref{tab:real-results} and visualized in Fig.~\ref{fig:real-charts}.

\begin{table*}
	\vspace{-1.5cm}
	\caption{Experiments with real data. Mean orientation error $\mathrm{\Delta\Theta}$ [°], position error $\mathrm{\Delta\Tau}$ [\,] and reprojection error $\mathrm{\Delta\pi}$ for each method and image sequence. The top-3 results for each sequence are typeset in bold and color-coded (\rtA{best}, \rtB{$2^\text{nd}$-best} and \rtC{$3^\text{rd}$-best} result).}
    \label{tab:real-results}
    \centering
   	\makebox[\textwidth][c]{
		\begin{tabular}{llrrrrrrrrrrrr}

	        \hline
	        \noalign{\smallskip}
            Dataset & & TFH & BB & STR & MH & COR & MC1 & MC2 & MC3 & ULB & WDC \\
            \hline
            \noalign{\smallskip}

            & $\mathrm{\Delta\Theta}$ & - & - & - & 4.96 & - & - & - & - & - & - \\
            Ansar & $\mathrm{\Delta\Tau}$ & - & - & - & 0.38 & - & - & - & - & - & - \\
            & $\mathrm{\Delta\pi}$ & - & - & - & 5e-05 & - & - & - & - & - & - \\
            \hline
            \noalign{\smallskip}

            & $\mathrm{\Delta\Theta}$ & 32.24 & 88.18 & 0.90 & 0.46 & 0.22 & 4.83 & 15.47 & 5.00 & 2.51 & 36.52 \\
            Mirzaei & $\mathrm{\Delta\Tau}$ & 11.04 & 168.47 & 1.92 & \rC{0.04} & 0.10 & 1.53 & 7.37 & 1.82 & 1.27 & 6.44 \\
            & $\mathrm{\Delta\pi}$ & 1e+06 & 2e+06 & 8e-07 & 4e-07 & 1e-06 & 3e-06 & 3e-05 & 1e-02 & 2e-06 & 7e+03  \\
            \hline
            \noalign{\smallskip}

            & $\mathrm{\Delta\Theta}$ & 20.46 & 23.27 & 4.91 & 0.61 & 0.40 & 1.45 & 0.43 & 2.33 & 3.96 & 0.50 \\
            RPnL & $\mathrm{\Delta\Tau}$ & 15.32 & 53.03 & 9.73 & 0.07 & 0.13 & 0.43 & 0.22 & 1.22 & 2.08 & 0.23 \\
            & $\mathrm{\Delta\pi}$ & 6e-05 & 7e-06 & 9e-05 & 3e-06 & 6e-06 & 2e-06 & 1e-07 & 2e-05 & 6e-06 & 1e-06 \\
            \hline
            \noalign{\smallskip}

            & $\mathrm{\Delta\Theta}$ & 7.76 & 37.82 & 22.08 & \rA{0.25} & \rA{0.10} & 0.15 & 0.20 & 2.08 & 4.89 & 0.51 \\
            ASPnL & $\mathrm{\Delta\Tau}$ & 6.11 & 76.61 & 30.47 & \rA{0.02} & \rA{0.03} & \rC{0.04} & 0.08 & 0.74 & 2.22 & 0.23 \\
            & $\mathrm{\Delta\pi}$ & 6e-04 & 2e+03 & 3e+02 & \rA{5e-08} & \rA{9e-08} & 2e-08 & 1e-08 & 4e-06 & 3e-06 & 1e-06 \\
            \hline
            \noalign{\smallskip}

            & $\mathrm{\Delta\Theta}$ & 1.10 & 1.98 & \rC{0.15} & 0.45 & 0.13 & \rA{0.03} & \rA{0.03} & \rB{0.09} & 0.49 & \rC{0.18} \\
            LPnL\_Bar\_LS & $\mathrm{\Delta\Tau}$ & 1.05 & \rC{7.23} & \rC{0.27} & \rC{0.04} & 0.05 & \rA{0.01} & \rA{0.02} & \rB{0.03} & 0.22 & \rC{0.11} \\
            & $\mathrm{\Delta\pi}$ & 7e-07 & 1e-06 & 8e-08 & 8e-07 & 1e-06 & \rB{2e-09} & \rB{1e-09} & \rB{6e-08} & 2e-07 & \rB{4e-08} \\
            \hline
            \noalign{\smallskip}

            & $\mathrm{\Delta\Theta}$ & \rC{0.57} & \rA{0.30} & \rA{0.11} & \rB{0.32} & \rA{0.10} & \rB{0.04} & \rA{0.03} & \rA{0.07} & \rC{0.39} & \rA{0.08} \\
            LPnL\_Bar\_ENull & $\mathrm{\Delta\Tau}$ & \rC{0.45} & \rA{1.13} & \rA{0.16} & \rA{0.02} & \rB{0.04} & \rA{0.01} & \rA{0.02} & \rA{0.02} & \rC{0.18} & \rA{0.05} \\
            & $\mathrm{\Delta\pi}$ & \rB{2e-07} & \rA{2e-08} & \rB{3e-08} & \rB{2e-07} & \rC{4e-07} & \rA{8e-10} & \rA{7e-10} & \rA{5e-08} & \rC{1e-07} & \rA{2e-08} \\
            \hline
            \noalign{\smallskip}

            & $\mathrm{\Delta\Theta}$ & \rB{0.47} & 2.18 & \rA{0.11} & 0.95 & 0.12 & 0.12 & 0.28 & 0.23 & \rB{0.23} & \rB{0.16} \\
            DLT-Lines & $\mathrm{\Delta\Tau}$ & \rB{0.44} & 8.11 & \rB{0.18} & 0.09 & 0.05 & \rC{0.04} & 0.16 & 0.08 & \rB{0.10} & \rB{0.10} \\
            & $\mathrm{\Delta\pi}$ & \rB{2e-07} & 1e-06 & \rA{2e-08} & 1e-06 & 2e-06 & \rC{6e-09} & 4e-08 & 3e-07 & \rA{3e-08} & \rC{6e-08} \\
            \hline
            \noalign{\smallskip}

            & $\mathrm{\Delta\Theta}$ & 1.11 & \rC{1.04} & 0.93 & 17.58 & 0.38 & 0.28 & 0.22 & 0.48 & 0.77 & 0.34 \\
            DLT-Plücker-Lines & $\mathrm{\Delta\Tau}$ & 1.28 & 11.69 & 1.78 & 0.74 & 0.13 & 0.40 & 0.50 & 0.27 & 0.47 & 0.39 \\
            & $\mathrm{\Delta\pi}$ & 1e-06 & \rC{8e-07} & 2e-06 & 3e-02 & 3e-06 & 2e-06 & 9e-07 & 2e-05 & 8e-07 & 1e-06 \\
            \hline
            \noalign{\smallskip}

            & $\mathrm{\Delta\Theta}$ & \rA{0.39} & \rB{0.40} & 0.22 & \rC{0.41} & \rC{0.11} & \rC{0.11} & \rC{0.15} & \rC{0.16} & \rA{0.20} & 0.23 \\
            DLT-Combined-Lines & $\mathrm{\Delta\Tau}$ & \rA{0.32} & \rB{1.88} & 0.38 & \rC{0.04} & \rB{0.04} & \rC{0.04} & \rC{0.07} & \rC{0.05} & \rA{0.08} & 0.12 \\
            & $\mathrm{\Delta\pi}$ & \rA{7e-08} & \rB{4e-08} & \rC{6e-08} & \rC{3e-07} & \rB{2e-07} & 2e-08 & \rC{2e-08} & \rC{2e-07} & \rB{7e-08} & 2e-07 \\
            \hline
		\end{tabular}
	}
\end{table*}

On sequences with small number of lines (MH 30, COR 69), the results of non-LPnL and LPnL methods are comparable. Conversely, on sequences with high number of lines (177 -- 1841), the non-LPnL methods are usually less accurate than the LPnL methods.
\textbf{Ansar}\,\markAnsar\ was run only on the MH sequence containing 30 lines, because other sequences caused it to exceed available memory. It achieves poor performance.
\textbf{Mirzaei}\,\markMirzaei\ yields usually the least accurate estimate on sequences with high number of lines.
On other sequences, it performs comparably to the other methods.
A slightly better accuracy is achieved by \textbf{RPnL}\,\markRPnL, but it also has trouble on sequences with high number of lines (TFH, BB, STR).
The related method \textbf{ASPnL}\,\markASPnL\ mostly performs better than \textbf{RPnL} with an exception of sequences with many lines -- BB and STR. Nevertheless, \textbf{ASPnL} yields the most accurate pose estimates on MH and COR. This agrees with the findings of \cite{xu2016pnl}, who state that \textbf{ASPnL} is suitable rather for small line sets.

The most accurate results on each sequence are predominantly achieved by the LPnL methods:
Most of the top-3 results are achieved by \textbf{LPnL\-\_Bar\-\_ENull}\,\markLPnLBarENull, followed by the proposed method \textbf{DLT-Combined-Lines}\,\markDLTcombined\,, see Table~\ref{tab:real-results}.
\textbf{LPnL\-\_Bar\-\_LS}\,\markLPnLBarLS\ and \textbf{DLT-Lines}\,\markDLTlines\ also achieve top-3 accuracy, although it happens less frequently.
\textbf{DLT-Plücker-Lines}\,\markDLTplucker\ is the least accurate LPnL method on real data, being the only LPnL method which performs slightly below expectations based on synthetic data.
Results of other methods are consistent with the results achieved on synthetic lines.

\subsection{Effect of prenormalization}
\label{subsec:results-prenormalization}

The prenormalization procedure of the proposed method was described in Section~\ref{subsec:DLT-combined-prenormalization}. It has five stages, which are labeled (i) -- (v).
To show how the individual stages contribute to the overall accuracy of the method, it was executed on both synthetic lines and real data, while the prenormalization stages were activated one after another.
The experimental setup for synthetic lines was the same as in Section~\ref{subsec:results-synthetic}.

The results are shown in Table~\ref{tab:prenorm-results}, which contains the median errors in estimated camera pose for each group of active stages. Each row also shows a relative improvement in accuracy with respect to the previous row.

\begin{table*}
	\caption{Effect of prenormalization stages on the accuracy of the proposed method.}
    \label{tab:prenorm-results}
    \centering
   	\makebox[\textwidth][c]{
		\begin{tabular}{l|rrrrrr|rrrrrr}

        & \multicolumn{6}{c|}{\underline{\smash{Synthetic lines}}} & \multicolumn{6}{c}{\underline{\smash{Real data}}} \\
        Prenorm. & \multicolumn{3}{c}{\underline{\smash{Median of abs. error}}} & \multicolumn{3}{c|}{\underline{\hspace{1em}\smash{Improvement}\hspace{1em}}} & \multicolumn{3}{c}{\underline{\smash{Median of abs. error}}} & \multicolumn{3}{c}{\underline{\hspace{1em}\smash{Improvement}\hspace{1em}}} \\
        stages & $\mathrm{\Delta\Theta}$\,[°] & \hspace{-0.5em}$\mathrm{\Delta\Tau}$\,[m] & $\mathrm{\Delta\pi}$\,[\ ] & $\mathrm{\Delta\Theta}$ & $\mathrm{\Delta\Tau}$ & $\mathrm{\Delta\pi}$  &  $\mathrm{\Delta\Theta}$\,[°] & \hspace{-0.5em}$\mathrm{\Delta\Tau}$\,[\ ] & $\mathrm{\Delta\pi}$\,[\ ] & $\mathrm{\Delta\Theta}$ & $\mathrm{\Delta\Tau}$ & $\mathrm{\Delta\pi}$ \\
        \hline
        none & 0.92 & 444.48 & 1.11e-2 & --- & --- & --- & 1.25 & 0.21 & 2.40e-6 & --- & --- & --- \\
        (i) & 0.93 & 373.35 & 1.50e-2 & -1\,\% & 16\,\% & -35\,\% & 0.38 & 0.14 & 1.42e-7 & 69\,\% & 34\,\% & 94\,\% \\
        (i), (ii) & 0.01 & 0.48 & 3.17e-6 & 98\,\% & 100\,\% & 100\,\% & 0.22 & 0.09 & 6.25e-8 & 42\,\% & 33\,\% & 56\,\% \\
        (i) -- (iii) & 0.01 & 0.48 & 3.15e-6 & 0\,\% & 0\,\% & 0\,\% & 0.22 & 0.09 & 6.26e-8 & 0\,\% & 0\,\% & 0\,\% \\
        (i) -- (iv) & 0.01 & 0.48 & 3.08e-6 & 0\,\% & 0\,\% & 2\,\% & 0.23 & 0.09 & 6.33e-8 & -1\,\% & 0\,\% & -1\,\% \\
        (i) -- (v) & 0.01 & 0.64 & 1.47e-6 & 2\,\% & -34\,\% & 52\,\% & 0.21 & 0.07 & 6.93e-8 & 8\,\% & 21\,\% & -9\,\% \\
		\end{tabular}
	}
\end{table*}

The measurements show that stages (i) and (ii), i.\,e.\ multiplication of each 3D point/line by a constant and translation of the data to be centered around the origin, have the biggest impact on accuracy of the method.
Stage (v), i.\,e.\ scaling of some parts of the measurement matrix, proves to be important mainly in the case of real data.
Stages (iii) and (iv), i.\,e.\ the second translation and anisotropic scaling, have a minor positive impact on accuracy.

\subsection{Speed}
\label{subsec:results-speed}

Efficiency of each method was evaluated by measuring runtime on a desktop PC with a quad core Intel i5 3.33\,GHz CPU and 10\,GB of RAM. The experimental setup was the same as in Section~\ref{subsec:results-synthetic}, varying the number of lines.

\begin{figure}
	\centering
	\includegraphics[width=0.7\linewidth]{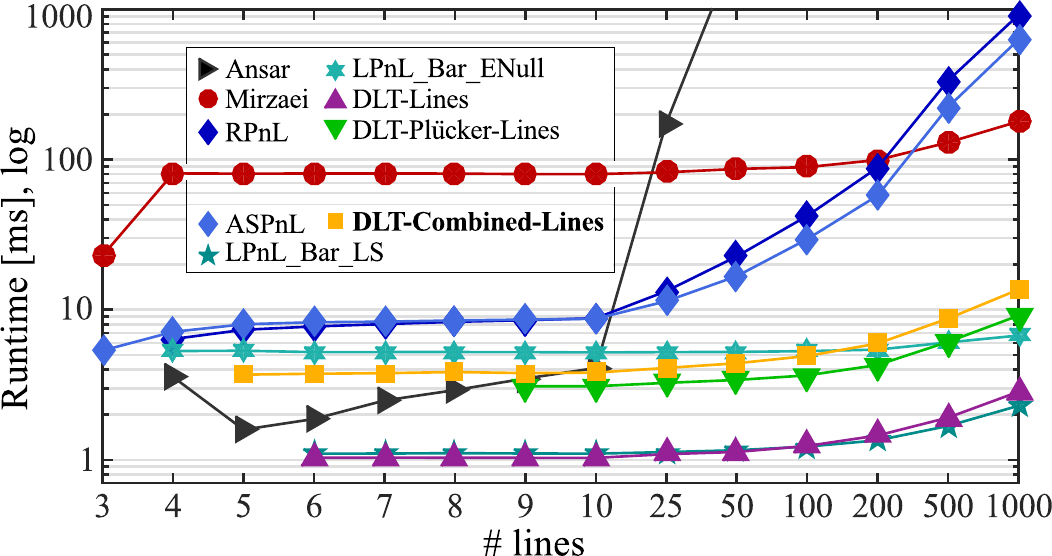}
    \caption{Runtimes as a function of the number of lines, averaged over 1000 trials. Logarithmic vertical axis.}
	\label{fig:runtime}
\end{figure}

As it can be seen in Figure~\ref{fig:runtime} and Table~\ref{tab:runtime}, the only method with $O(m^2)$ computational complexity in the number of lines $m$ is \textbf{Ansar}\,\markAnsar.
The space complexity of the implementation used is apparently also quadratic. We were unable to execute it for as few as 100 lines due to lack of computer memory.
All other tested methods have $O(m)$ computational complexity.
The runtimes however differ substantially. It is apparent that the LPnL methods are significantly faster than the non-LPnL methods.

\begin{table}
	\caption{Runtimes in milliseconds for varying number of lines, averaged over 1000 trials.}
    \label{tab:runtime}
    \centering
	\begin{tabular}{lrrrr}
		\hline\noalign{\smallskip}
		\#\,lines & 10 & 100 & 1000\\
		\noalign{\smallskip}
		\hline
		\noalign{\smallskip}
   		Ansar  & 4.1 & - & -\\
		Mirzaei  & 77.9 & 84.2 & 155.2\\
		RPnL & 8.8 & 41.3 & 879.5\\
        ASPnL  & 8.7 & 29.5 & 630.2\\
        LPnL\_Bar\_LS  & 1.1 & 1.2 & 2.3\\
        LPnL\_Bar\_ENull  & 5.2 & 5.3 & 6.7\\
   		DLT-Lines & 1.0 & 1.2 & 2.7\\
   		DLT-Plücker-Lines & 3.0 & 3.6 & 8.2\\
		DLT-Combined-Lines & 3.7 & 4.6 & 12.1\\
		\hline
	\end{tabular}
\end{table}

\textbf{RPnL}\,\markRPnL\ and \textbf{ASPnL}\,\markASPnL, being related methods, are nearly equally fast. \textbf{RPnL} is slightly faster for $m < 10$ lines, while \textbf{ASPnL} is faster for greater numbers of lines. Runtimes of both methods rise steeply with increasing number of lines, reaching 630.2\,ms on 1000 lines for \textbf{ASPnL}.
Runtime of \textbf{Mirzaei}\,\markMirzaei, on the other hand, grows very slowly, spending 155.2\,ms on 1000 lines.
However, \textbf{Mirzaei} is slower than \textbf{RPnL} for $m < 200$ lines.
This is due to computation of a $120 \times 120$ Macaulay matrix in Mirzaei's method which has an effect of a constant time penalty.

The LPnL methods are one to two orders of magnitude faster than the non-LPnL methods. The fastest two are \textbf{DLT-Lines}\,\markDLTlines\ and \textbf{LPnL\-\_Bar\-\_LS}\,\markLPnLBarLS, spending about 1\,ms on 10 lines, and not more than 3\,ms on 1000 lines, see Table~\ref{tab:runtime}.
Slightly slower are \textbf{DLT-Plücker-Lines}\,\markDLTplucker, \textbf{DLT-Combined-Lines}\,\markDLTcombined\ and \textbf{LPnL\_Bar\_ENull}\,\markLPnLBarENull, spending about 3 -- 5\,ms on 10 lines, and about 6 -- 12\,ms on 1000 lines.
The slowdown factor for \textbf{DLT-Plücker-Lines} is the prenormalization of 3D lines.
This is also the case of \textbf{DLT-Combined-Lines}, where a measurement matrix of double size must be additionally decomposed compared to the competing methods, see Eq.~(\ref{eq:combinedmeasurementmatrix-stacked}).
The computationally demanding part of \textbf{LPnL\-\_Bar\-\_ENull} is the effective null space solver carrying out Gauss-Newton optimization.

\subsection{Robustness to outliers}
\label{subsec:results-outliers}

As a practical requirement, robustness to outlying correspondences was also tested.
The experimental setup was the same as in Section~\ref{subsec:results-synthetic}, using $m = 500$ lines having endpoints perturbed with slight image noise ($\sigma = 2$ pixels).
The image lines simulating outlying correspondences were perturbed with an additional extreme noise with $\sigma = 100$ pixels.
The fraction of outliers varied from 0\,\% to 80\,\%.

\textbf{Ansar}, \textbf{Mirzaei}, and \textbf{RPnL} methods were plugged into a MLESAC framework \cite[a generalization of RANSAC which maximizes the likelihood rather than just the number of inliers]{torr2000mlesac}.
Since \textbf{Ansar} cannot handle the final pose computation from potentially hundreds of inlying line correspondences, it is computed by \textbf{RPnL}.
The probability that only inliers will be selected in some iteration was set to 99\,\%, and the number of iterations was limited to 10,000.
The inlying correspondences were identified based on the line reprojection error.
No heuristic for early hypothesis rejection was utilized, as it can also be incorporated into AOR, e.\,g.\ by weighting the line correspondences.
\textbf{DLT-Lines}, \textbf{DLT-Plücker-Lines}, and \textbf{DLT-Combined-Lines} methods were equipped with AOR, which was set up as described in Section~\ref{subsec:AOR}.

The setup presented by \cite{xu2016pnl} was also tested: \textbf{LPnL\-\_Bar\-\_LS} and \textbf{LPnL\-\_Bar\-\_ENull} methods with AOR, and a \textbf{P3L} solver and \textbf{ASPnL} plugged into a RANSAC framework, generating camera pose hypotheses from 3 and 4 lines, respectively.
The authors have set the required number of inlying correspondences to 40\,\% of all correspondences, and limit the number of iterations to 80.
When this is exceeded, the required number of inliers is decreased by a factor of 0.5, and another 80 iterations are allowed.
The inlying correspondences are identified based on thresholding of an algebraic error -- the residuals $\epsilon_i$ of the least squares solution in Eq.~(\ref{eq:noisysystem}), where the measurement matrix $\dot{\mathsf{M}}$ is used, defined by Eq.~(\ref{eq:pointmeasurementmatrix}).

The tested methods are summarized in the following list (the number at the end of MLE\-SAC/RAN\-SAC denotes the number of lines used to generate hypotheses).

\begin{enumerate}
	\item \textbf{Ansar\,+\,MLESAC4\,+\,RPnL}, Ansar plugged into a MLESAC loop, the final solution computed by RPnL. Results shown in black \markAnsar\,.
	\item \textbf{Mirzaei\,+\,MLESAC3}, results shown in red \markMirzaei\,.
	\item \textbf{RPnL\,+\,MLESAC4}, results shown in blue \markRPnL\,.
    \item \textbf{P3L\,+\,RANSAC3}, the setup by \cite{xu2016pnl}, results shown in sky blue \markPthreeL\,.
    \item \textbf{ASPnL\,+\,RANSAC4}, the setup by \cite{xu2016pnl}, results shown in  light blue \markASPnL\,.
    \item \textbf{LPnL\-\_Bar\-\_LS\,+\,AOR}, the setup by \cite{xu2016pnl}, results shown in teal \markLPnLBarLS\,.
    \item \textbf{LPnL\-\_Bar\-\_ENull\,+\,AOR}, the setup by \cite{xu2016pnl}, results shown in blue-green \markLPnLBarENull\,.
    \item \textbf{DLT-Lines\,+\,AOR}, results shown in purple \markDLTlines\,.
	\item \textbf{DLT-Plücker-Lines\,+\,AOR}, results shown in green \markDLTplucker.
    \item \textbf{DLT-Combined-Lines\,+\,AOR}, the proposed method with AOR, results shown in orange \markDLTcombined\,.
\end{enumerate}

The RANSAC-based approaches can theoretically handle any percentage of outliers.
This is confirmed by \textbf{Mirzaei\,+\,MLE\-SAC3}\,\markMirzaei\ and \textbf{RPnL\,+\,MLE\-SAC4}\,\markRPnL, as their accuracy does not change with respect to the fraction of outliers.
What does change however, is the number of iterations (and thus also the runtime).
Even then, the limit of 10,000 iterations was almost never reached.
A different situation occurred when testing \textbf{Ansar\,+\,MLE\-SAC3\,+\,RPnL}\,\markAnsar, where the iteration limit was sometimes reached even at 20\,\% of outliers. 
This suggests that \textbf{Ansar} is a poor hypothesis generator, and the MLESAC framework needs to iterate more times to get a valid hypothesis.

\begin{figure*}
	\centering
		\makebox[\textwidth][c]{\includegraphics[width=1.4\linewidth]{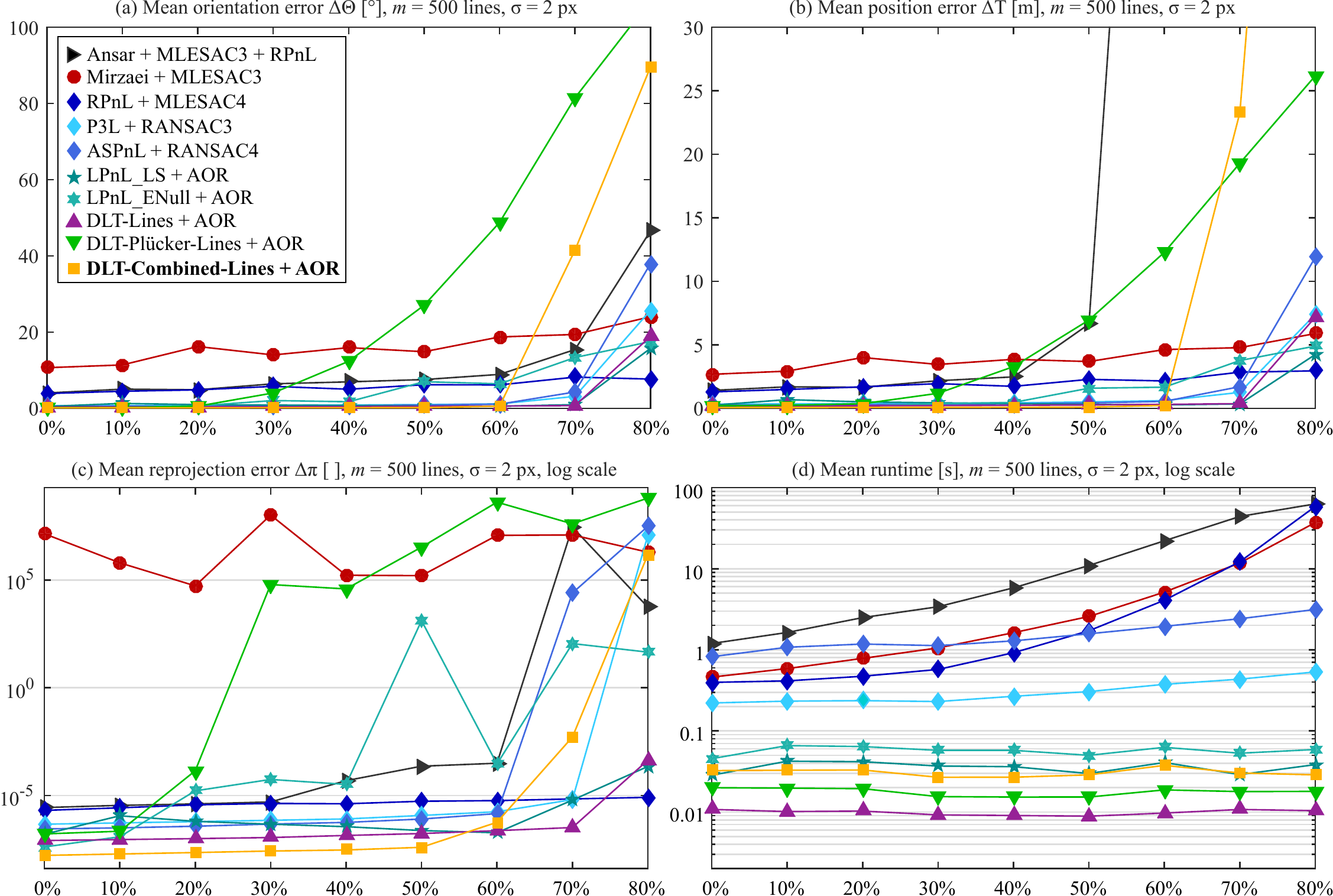}}
    \caption{
		Experiments with outliers. Mean camera orientation errors ($\mathrm{\Delta\Theta}$,~\emph{a}), position errors ($\mathrm{\Delta\Tau}$,~\emph{b}), reprojection errors ($\mathrm{\Delta\pi}$,~\emph{c}) and runtimes~(\emph{d}) depending on the percentage of outliers. Averaged over 1000 trials.
	}
	\label{fig:outliers}
\end{figure*}

\textbf{P3L\,+\,RAN\-SAC3}\,\markPthreeL\ and \textbf{ASPnL\,+\,RAN\-SAC4}\,\markASPnL\ have much lower runtimes, which is caused mainly by the setup limiting the number of iterations to a few hundreds.
The setup has, on the other hand, a negative effect on the robustness of the method: the break-down point is only 60 -- 70\,\%, as it is apparent in Figs.~\ref{fig:outliers}a, b.
This issue was not observed by \cite{xu2016pnl}, because they tested the methods only up to 60\,\% of outliers.

LPnL methods with AOR have constant runtimes regardless of the fraction of outliers. 
The fastest one is \textbf{DLT-Lines\,+\,AOR}\,\markDLTlines\ running in 10\,ms on average.
The proposed method \textbf{DLT-Combined-Lines\,+\,AOR}\,\markDLTcombined\ runs in 31\,ms on average, and \textbf{LPnL\-\_Bar\-\_ENull\,+\,AOR}\,\markLPnLBarENull\ is the slowest one with 57\,ms, see Fig.~\ref{fig:outliers}d.
The robustness of the LPnL methods differs significantly.
\textbf{DLT-Plücker-Lines\,+\,AOR}\,\markDLTplucker\ breaks-down at about 40\,\%, but it occasionally generates wrong solutions from 30\,\% up, 
We call this a ``soft'' break-down point.
\textbf{LPnL\-\_Bar\-\_ENull\,+\,AOR}\,\markLPnLBarENull\ behaves similarly, but it yields smaller pose errors.
\textbf{DLT-Lines\,+\,AOR}\,\markDLTlines\,, \textbf{LPnL\-\_Bar\-\_LS\,+\,AOR}\,\markLPnLBarLS, and the proposed method \textbf{DLT-Combined-Lines\,+\,AOR}\,\markDLTcombined\,, on the other hand, have a ``hard'' break-down point at 70\,\%, 65\,\%, and 60\,\%, respectively. This means they do not yield wrong solutions until they reach the break-down point.

The RANSAC-based approach is irreplaceable in cases with high percentage of outliers. Aside from this, for lower fractions of outliers, the LPnL\,+\,AOR alternatives are more accurate and 4 -- 31$\times$ faster than the RANSAC-based approaches, depending on the chosen LPnL method.

The distributions of errors of the tested methods over all 1000 trials are provided in the supplementary material.

\section{Conclusions}
\label{sec:conclusions}

A novel algebraic method DLT-Combined-Lines is proposed to estimate camera pose from line correspondences. The method is based on linear formulation of the Perspective-n-Line problem, and it uses Direct Linear Transformation to recover the combined projection matrix.
The matrix is a combination of projection matrices used by the DLT-Lines and DLT-Plücker-Lines methods, that work with 3D points and 3D lines, respectively. The proposed method works with both 3D points and lines, leading to the reduction of the minimum of required lines to 5, and it can be easily extended to use 2D points as well.
The combined projection matrix contains multiple estimates of camera rotation and translation, which can be recovered after enforcing constraints of the matrix. Multiplicity of the estimates leads to better accuracy compared to the other DLT methods.

For small line sets, the proposed method is not as accurate as the non-LPnL methods, which is a common attribute of all methods using linear formulation. For larger line sets, the method is comparable to the state-of-the-art method LPnL\-\_Bar\-\_ENull in accuracy of orientation estimation. Yet, it is more accurate in estimation of camera position and it yields smaller reprojection error under strong image noise. On real world data, the proposed method achieves top-3 results.
It also keeps the common advantage of LPnL methods: very high computational efficiency.
To deal with outliers, the proposed method is equipped with the Algebraic Outlier Rejection scheme,
being able to handle up to 65\,\% of outliers in a fraction of time required by the RANSAC-based approaches.

It is clear that none of the existing methods is universally best. We thus suggest to use \textbf{ASPnL} for small line sets ($m \le 10$).
For bigger line sets ($m > 10$), we suggest to use a LPnL method: either \textbf{LPnL\-\_Bar\-\_ENull} (in cases with small noise or quasi-singular line configurations) or \textbf{DLT-Combined-Lines} (in cases with many lines and/or strong noise).

Future work involves examination of the combined projection matrix in order to adaptively combine the multiple camera rotation and translation estimates contained in the matrix.
Inspired by the work of \citeauthor{xu2016pnl}, the proposed method could also be combined with the effective null space solver. This might further increase the accuracy of the method.

Matlab code of the proposed method and the supplementary material are publicly available at \url{\codeURL}.

\appendix
\section{Derivation of a measurement matrix $\mathsf{M}$ from 3D/2D correspondences}
\label{sec:appendix-corresp}

Correspondences between 3D entities and their 2D counterparts are defined by equations which, in turn, generate rows of a measurement matrix $\mathsf{M}$. The following derivations are made for a single 3D/2D correspondence.
More correspondences lead simply to stacking the rows of $\mathsf{M}$.

\subsection{Line-line correspondence}
\label{subsec:appendix-corresp-line-line}

We start from Eq.~(\ref{eq:lineprojection}) defining the projection of a 3D line $\mathbf{L}$ by a line projection matrix $\bar{\mathsf{P}}$ onto the image line $\mathbf{l}$
\begin{equation}
	\mathbf{l} \approx \bar{\mathsf{P}} \mathbf{L} \enspace .
\end{equation}

\noindent We swap its sides and pre-multiply them by $[\mathbf{l}]_\times$
\begin{equation}
	[\mathbf{l}]_\times \bar{\mathsf{P}} \mathbf{L} \approx [\mathbf{l}]_\times \mathbf{l} \enspace .
\end{equation}

\noindent The right-hand side is apparently a vector of zeros
\begin{equation}
    [\mathbf{l}]_\times \bar{\mathsf{P}} \mathbf{L} = \mathbf{0} \enspace .
\end{equation}

\noindent Using Lemma 4.3.1 of \cite{horn1994topics}, we get
\begin{equation}
    \left( \mathbf{L}^\top \otimes [\mathbf{l}]_\times \right) \cdot \mathrm{vec}(\bar{\mathsf{P}}) = \mathbf{0} \enspace .
\end{equation}

\noindent The left-hand side can be divided into the measurement matrix $\mathsf{M} = \mathbf{L}^\top \otimes [\mathbf{l}]_\times$ and the vector of unknowns $\bar{\mathbf{p}} = \mathrm{vec}(\bar{\mathsf{P}})$, finally yielding the homogeneous system
\begin{equation}
    \mathsf{M} \bar{\mathbf{p}} = \mathbf{0} \enspace .
\end{equation}

\subsection{Point-point correspondence}
\label{subsec:appendix-corresp-point-point}

The derivation is the same as in the case of line-line correspondences, but starting from Eq.~(\ref{eq:pointprojection}) defining the projection of a 3D point $\mathbf{X}$ by a point projection matrix $\dot{\mathsf{P}}$ onto the image point $\mathbf{x}$.
\begin{align}
	\mathbf{x} &\approx \dot{\mathsf{P}} \mathbf{X} \\
	[\mathbf{x}]_\times \dot{\mathsf{P}} \mathbf{X} &\approx [\mathbf{x}]_\times \mathbf{x} \\
    [\mathbf{x}]_\times \dot{\mathsf{P}} \mathbf{X} &= \mathbf{0} \\
    \left( \mathbf{X}^\top \otimes [\mathbf{x}]_\times \right) \cdot \mathrm{vec}(\dot{\mathsf{P}}) &= \mathbf{0} \\
    \mathsf{M} \dot{\mathbf{p}} &= \mathbf{0}
\end{align}

\subsection{Point-line correspondence}
\label{subsec:appendix-corresp-point-line}

We start from Eq.~(\ref{eq:DLT-Lines}) relating the projection of a 3D point $\mathbf{X}$ and an image line $\mathbf{l}$
\begin{equation}
	\label{eq:point-line-corresp}
    \mathbf{l}^\top \dot{\mathsf{P}} \mathbf{X} = 0 \enspace .
\end{equation}

\noindent Since Eq.~(\ref{eq:point-line-corresp}) already has the right-hand side equal to 0, we can directly apply Lemma 4.3.1 of \cite{horn1994topics}, and see how the measurement matrix $\mathsf{M}$ is generated:
\begin{align}
    \left( \mathbf{X}^\top \otimes \mathbf{l}^\top \right) \cdot \mathrm{vec}(\dot{\mathsf{P}}) &= 0 \enspace , \\
    \mathsf{M} \dot{\mathbf{p}} &= 0 \enspace .
\end{align}

\section*{Acknowledgements}
This work was supported by The Ministry of Education, Youth and Sports of the Czech Republic from the National Programme of Sustainability (NPU II); project IT4Innovations excellence in science -- LQ1602.
The authors would like to thank Tomáš Werner for proofreading of the paper.

\clearpage
\bibliography{references}

\end{document}